%% file: main.tex
\documentclass{article}

\usepackage[preprint]{icml2026}

\usepackage{microtype}
\usepackage{graphicx}
\usepackage{subcaption}
\usepackage{booktabs}
\usepackage{amsmath,amssymb,mathtools}
\usepackage{hyperref}
\usepackage[capitalize,noabbrev]{cleveref}
\usepackage{multirow}
\usepackage{makecell} 
\usepackage{placeins} % 在导言区添加
\usepackage[textsize=tiny]{todonotes}
\usepackage{algorithm}
\usepackage{algorithmic}
\icmltitlerunning{DiMo: Discrete Diffusion Modeling for Motion Generation and Understanding}

\begin{document}

\twocolumn[
\icmltitle{DiMo: Discrete Diffusion Modeling for Motion Generation and Understanding}
\icmlsetsymbol{equal}{*}

\begin{icmlauthorlist}
  \icmlauthor{Ning Zhang}{huawei1}
  \icmlauthor{Zhengyu Li}{huawei1}
  \icmlauthor{Kwong Weng Loh}{huawei1}
  \icmlauthor{Mingxi Xu}{huawei1}
  \icmlauthor{Qi Wang}{huawei1}
  \icmlauthor{Zhengyu Wen}{huawei1}
  \icmlauthor{Xiaoyu He}{huawei1}
  \icmlauthor{Wei Zhao}{huawei1}
  \icmlauthor{Kehong Gong}{huawei2}
  \icmlauthor{Mingyuan Zhang}{huawei1}
\end{icmlauthorlist}

\icmlaffiliation{huawei1}{Huawei Central Media Technology Institute}
\icmlaffiliation{huawei2}{Huawei Technologies Co., Ltd.} 
\icmlcorrespondingauthor{Mingyuan Zhang}{zhangmy718@gmail.com}

  % You may provide any keywords that you find helpful for describing your
  % paper; these are used to populate the "keywords" metadata in the PDF but
  % will not be shown in the document
  \icmlkeywords{Machine Learning, ICML}
  \vskip 0.2in 

%   % ======= 简化Teaser插入 =======
%   \centering
%   \icmlkeywords{Machine Learning, ICML}
 
% % \vspace{-6mm}

% \noindent
% \parbox{\textwidth}{
%     \centering
%     \includegraphics[width=0.95\textwidth]{Assets/images/teaser.png}
%     \vspace{6pt}
%     {\small
%     \emph{Figure 1.} Overview of DiMo. Our method achieves a better balance
%     on Text-to-Motion and Motion-to-Text tasks.
%     } 
% }
% \setcounter{figure}{1}
% \vspace{2mm}
  
]
\printAffiliationsAndNotice{}

\vspace{-6mm}
\input{sections/0_abstract}

\input{sections/1_introduction}

\input{sections/2_related_works}
\input{sections/4_method}
\input{sections/5_experiments}

\input{sections/6_applications}
\input{sections/7_conclusions}

% ================= References =================
\bibliography{icml2026}
\bibliographystyle{icml2026}
% In the unusual situation where you want a paper to appear in the
% references without citing it in the main text, use \nocite

% ================= Appendix =================
\newpage
\appendix
\onecolumn
\input{sections/8_appendix}

\end{document}

%% file: sections/0_abstract.tex
\begin{abstract}  

Prior masked modeling motion generation methods predominantly study text-to-motion. We present DiMo, a discrete diffusion-style framework, which extends masked modeling to bidirectional text--motion understanding and generation. Unlike GPT-style autoregressive approaches that tokenize motion and decode sequentially, DiMo performs iterative masked token refinement, unifying Text-to-Motion (T2M), Motion-to-Text (M2T), and text-free Motion-to-Motion (M2M) within a single model. This decoding paradigm naturally enables a quality-latency
trade-off at inference via the number of refinement steps.We further improve motion token fidelity with residual vector quantization (RVQ) and enhance alignment and controllability with Group Relative Policy Optimization (GRPO). Experiments on HumanML3D and KIT-ML show strong motion quality and competitive bidirectional understanding under a unified framework. In addition, we demonstrate model ability in text-free motion completion, text-guided motion prediction and motion caption correction without architectural change. Additional qualitative results are available on our project page:
\url{https://animotionlab.github.io/DiMo/}.

% We present DiMo, a discrete diffusion framework for bidirectional text-motion understanding and generation. Unlike GPT-style autoregressive approaches that tokenize motion and decode sequentially, DiMo performs multi-step parallel denoising, unifying Text-to-Motion (T2M), Motion-to-Text (M2T), and text-free Motion-to-Motion (M2M) within a single model. This decoding paradigm naturally enables a quality-latency trade-off at inference. On HumanML3D, our method achieves competitive T2M/M2T results against strong baselines.  Besides T2M/M2T, we further demonstrate motion completion and prediction under both text-conditioned and text-free settings. 
% We also incorporate Residual VQ (RVQ) as the motion tokenizer to improve quantization fidelity, and adopt Group Relative Policy Optimization (GRPO) within the framework to enhance alignment and controllability. To the best of our knowledge, this is the first work to bring diffusion-LLMs to bidirectional text-motion modeling.
\end{abstract}

%% file: sections/1_introduction.tex
\section{Introduction}

Bidirectional text–motion understanding and generation is increasingly important for modern motion systems~\cite{guo2022generating}. Complementing text2motion (T2M) and motion2text (M2T), many practical tasks such as caption correction, motion continuation, and motion in-betweening require a model to both interpret motion and generate it. Treating T2M and M2T as separate problems leads to inconsistent representations and duplicated engineering, whereas a unified framework provides shared cross-modal embeddings and consistent behavior across tasks. However, achieving high-quality generation, low-latency decoding, editability, and multi-task unification within a single model remains challenging.

Most prevailing approaches follow the autoregressive (AR) paradigm: continuous motions are discretized into tokens and decoded sequentially in a GPT-style manner to accomplish Text-to-Motion (T2M) and Motion-to-Text (M2T)~\cite{jiang2024motiongpt, wang2024motiongpt2, guo2022tm2t}. While this line of work has advanced unified interfaces and transferability, it faces inherent limitations: (i) token/frame-by-token decoding makes end-to-end latency scale with sequence length; (ii) the unidirectional nature of AR decoding hinders global, multi-step correction, which is important for editing and completion; and (iii) unifying text-conditioned and text-free variants of completion and prediction typically requires additional engineering branches and specialized training tricks. Recent module decoupling and cross-modal attention designs alleviate the tug-of-war between language and motion to some extent~\cite{zhu2025motiongpt3}, but decoding and editability remain fundamental bottlenecks.
\begin{figure}[t] 
\centering
\includegraphics[width=0.45\textwidth]{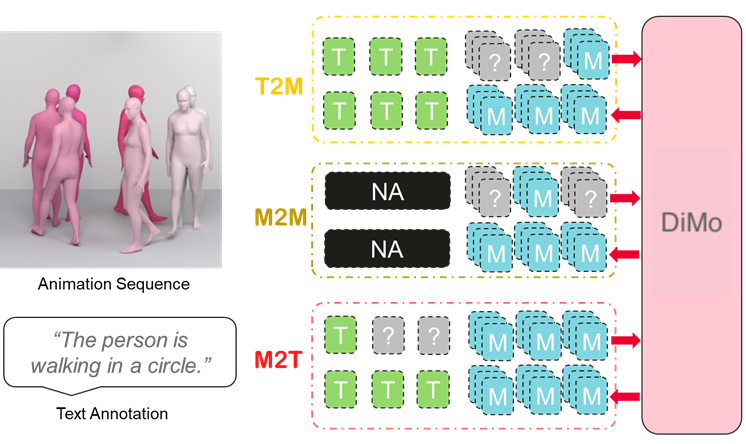}
\caption{Overview of DiMo. DiMo unifies Motion-to-Text (M2T) and Text-to-Motion (T2M) within a single framework, achieving a strong balance between motion realism and semantic consistency across generation and understanding tasks.}
\vspace{-5pt}
\label{fig:DiMo}
\end{figure}
We present \textbf{DiMo}, a \emph{discrete diffusion} framework for \emph{bidirectional} text--motion modeling. The key idea is to treat both text and motion as noisy sequences and perform K-step parallel denoising that progressively converges under global context. Unlike AR decoding, diffusion-style parallel denoising naturally supports parallel generation and multi-step self-correction, enabling a single model to unify T2M, M2T, and text-free Motion-to-Motion (M2M). Crucially, by adjusting the number of denoising steps at inference, the framework provides a tunable quality--latency trade-off, allowing practitioners to select Pareto-optimal operating points for different applications~\cite{li2022diffusionlm, nie2025llada}. To improve motion representation fidelity, we adopt Residual Vector Quantization (RVQ)~\cite{zeghidour2021soundstream} as the motion tokenizer, achieving lower quantization error at comparable bitrates; moreover, we further incorporate GRPO within the framework~\cite{shao2024deepseekmath} to enhance alignment and controllability. 
% The same paradigm naturally covers text-conditioned and text-free completion, prediction, and interpolation on the motion side, as well as motion-guided text correction and placeholder disambiguation on the M2T side.

On HumanML3D~\cite{guo2022generating}, our approach attains competitive results against strong baselines for both T2M and M2T, and empirically validates the quality--latency tunability of diffusion decoding.
% together with favorable scaling behavior: when the language backbone scales from \emph{ALBERT} $\rightarrow$ \emph{BERT-base} $\rightarrow$ \emph{BERT-large}~\cite{devlin2018bert}, performance consistently improves.
We also prototype text-free and text-conditioned motion completion and prediction, 
%and demonstrate motion-guided text correction and placeholder classification on the M2T side, further 
showcasing the unification and extensibility of DiMo.

Our contributions are threefold:
\begin{enumerate}
    \item \textbf{Paradigm \& tunable decoding.} We introduce the diffusion-LLMs training mechanism to bidirectional text--motion (T2M$\leftrightarrow$M2T) generation and propose a multi-step parallel denoising decoder under a unified framework. This paradigm naturally enables a quality--latency trade-off, which we validate with Pareto curves from step-count sweeps. Unlike strictly left-to-right decoding, iterative masked refinement enables step-by-step self-correction,
which empirically reduces sensitivity to early token mistakes and improves long-horizon coherence.
    \item \textbf{Unified capability \& natural extensibility.} Beyond standard T2M/M2T, our proposed framework can inherently support motion completion and prediction both text-free settings and text-conditioned on the T2M side, evidencing the paradigm's natural extensibility. On the M2T side, we prototype motion-guided text correction and placeholder disambiguation, demonstrating the robustness of our unified semantic space.
    \item \textbf{Practical effectiveness \& scalability.} We employ RVQ as the motion tokenizer to reduce quantization error and improve downstream quality, and integrate GRPO to enhance alignment and controllability; In HumanML3D and KIT-ML datasets, we achieve competitive results on both T2M and M2T tasks.
    % and observe consistent gains when scaling the language backbone from ALBERT $\rightarrow$ BERT-base $\rightarrow$ BERT-large.
\end{enumerate}

%% file: sections/2_related_works.tex
\section{Related Works}

% \begin{figure*}[t]
% \centering
% \includegraphics[width=0.85\textwidth]{Assets/images/teaser.png}
% \setlength{\abovecaptionskip}{0pt} % 控制 caption 上方间距（默认一般是 10pt 左右）
% \setlength{\belowcaptionskip}{0pt} % 控制 caption 下方间距
% \caption{Overview of DiMo. Our method achieves a better balance on T2M/M2T tasks.}
% % \vspace{-6pt}
% \label{fig:teaser}
% \end{figure*}

% \setlength{\dbltextfloatsep}{10pt}
% \begin{figure}[t]
%   % \vskip 0.2in
%   \begin{center}
%   \centerline{\includegraphics[width=0.95\columnwidth]{Assets/images/Picture4.jpg}}
%   \setlength{\abovecaptionskip}{0pt} % 控制 caption 上方间距（默认一般是 10pt 左右）
%   \setlength{\belowcaptionskip}{0pt} % 控制 caption 下方间距
%     \caption{
%       Overview of DiMo. Our method achieves a better balance on T2M/M2T tasks.
%     }
%     \vspace{-20pt}
%     \label{fig:teaser}
%   \end{center}
% \end{figure}

\subsection{Human Motion Generation}
Human motion generation has been a long-standing research problem at the intersection of computer vision, computer graphics, and machine learning. 
Early works relied on motion graphs and statistical models \citep{rose1998verbs, mukai2005geostatistical}, which enabled interpolation of motion clips but lacked semantic controllability. 
With the development of deep learning, generative models are introduced into this field. 
Generative adversarial networks (GANs) \citep{harvey2020robust, barsoum2018hp}, variational autoencoders (VAEs) \citep{aliakbarian2020stochastic, petrovich2021action}, and diffusion-based approaches \citep{tevet2022human, zhang2024motiondiffuse} further advanced the field, improving the diversity and realism of synthesized sequences. Later works also focus on higher motion controllability~\citep{zhang2024finemogen, karunratanakul2023guided}, motion generation under multimodal control signal~\citep{gong2023tm2d,zhang2024large}.

\subsection{Bi-directional Motion-Text Generation}
Bridging motion and natural language has attracted increasing interest for applications such as video captioning, embodied AI, and human-robot interaction. 
Text-to-motion (T2M) models \citep{guo2022generating, petrovich2022temos, zhang2024motiondiffuse} synthesize 3D motion sequences aligned with textual prompts. 
A notable effort toward unification is MotionGPT \citep{jiang2024motiongpt}, which quantizes human motion into discrete tokens and treats them as part of a shared vocabulary with language, enabling both T2M and M2T through a large language model backbone. 
While MotionGPT demonstrates strong bi-directional performance, its autoregressive generation can be inefficient for long motion sequences and may struggle to capture structural refinements. 
MotionGPT-2 \citep{wang2024motiongpt2} extends this line by integrating multimodal controls such as single-frame poses into the same framework. 
Other bi-directional approaches, such as TM2T~\citep{guo2022tm2t}, typically train separate models for each direction, lacking a unified architecture. 
Our work differs by combining the bi-directional modeling capacity of MotionGPT with the efficiency of masked refinement.

\subsection{Diffusion Language Models (dLLMs)}

While autoregressive (AR) large language models (LLMs) such as GPT-style transformers \citep{radford2018improving,brown2020language} dominate current research and applications, an alternative line of work explores bidirectional denoising-based models.  
This family, often referred to as diffusion language models (dLLMs), extends the principles of discrete diffusion \citep{austin2021structured, nie2025llada} and masked language modeling \citep{devlin2018bert} to sequential text generation.  

Instead of predicting tokens strictly left-to-right, dLLMs apply a diffusion-based noise schedule that progressively corrupts token sequences, and then learn to iteratively denoise them.  
Representative works include Diffusion-LM \citep{li2022diffusionlm}, which formulates text generation as discrete denoising diffusion, and MaskGIT \citep{chang2022maskgit}, which introduces iterative masked prediction for efficient parallel decoding.  
Recent surveys \citep{yu2025discrete} highlight that this paradigm achieves competitive quality compared to AR models, while offering controllable sampling, parallel decoding, and a natural quality–latency trade-off.  

Building on these insights, our proposed DiMo adapts the dLLM paradigm to motion--language modeling, integrating residual vector quantization for motion tokenization with diffusion-style denoising to support bidirectional generation across text and motion. For the language backbone, we adopt the BERT family, whose bidirectional masked language modeling naturally fits our denoising framework and is well aligned with the size of available motion–language datasets.

\label{sec:related_works}

%% file: sections/4_method.tex
\section{Methods}
\label{sec:methods}

\setlength{\dbltextfloatsep}{10pt}

\subsection{Overview}
We propose \textbf{DiMo}, a unified framework for bidirectional text-motion generation, covering text-to-motion (T2M), motion-to-text (M2T), and tasks such as motion completion and prediction. DiMo formulates both text and motion as discrete token sequences, applies a diffusion-style corruption process, and trains the model to iteratively denoise them, enabling multi-step parallel decoding that balances generation quality and latency.

Our design combines three components: (i) a Residual Vector Quantizer (RVQ) for high-fidelity motion tokens; (ii) a BERT-based masked language model backbone with a motion encoder-decoder, which fuses motion and text embeddings in a shared denoising framework; and (iii) an optional GRPO-based reinforcement objective for improved cross-modal alignment. This allows DiMo to achieve flexible and high-quality motion-language modeling.

\subsection{DiMo}
\label{sec:DiMo}
DiMo is a unified framework for bidirectional text-motion generation, building on the recent success of discrete diffusion language models (dLLMs). Instead of sequentially autoregressing tokens, dLLMs apply random masking and iterative denoising, which naturally supports parallel inference. This allows the model to refine corrupted sequences in multiple steps, dynamically revise low-confidence predictions, and leverage bidirectional attention for stronger contextual reasoning.

% \paragraph{Why Motion Fits Masked Modeling.}
% Motion data is inherently spatiotemporal: each frame depends on local kinematics while remaining globally constrained by long-range dynamics. Masked denoising is particularly suitable here, as it allows the model to recover missing trajectories using surrounding context, while iterative refinement prevents error accumulation over long horizons. This makes motion a natural candidate for masked generation, analogous to text but with temporal continuity as an additional inductive bias.

\paragraph{Multi-task Scheduling.}
To unify multiple objectives, DiMo employs a multi-task scheduling mechanism. During training, each sample within a batch is randomly assigned to one of three tasks: text-to-motion (T2M), motion-to-text (M2T), or motion-to-motion (M2M). The proportion of tasks is controlled by a tunable hyperparameter, enabling fine-grained adjustment of cross-modal versus uni-modal supervision. The definitions of these three tasks are listed as below:
\begin{itemize}
\item \textbf{Text-to-Motion (T2M):} The model learns to recover corrupted motion sequences from an unmasked text prompt by reconstructing masked motion tokens. All text tokens are reserved while motion tokens are randomly masked out.
\item \textbf{Motion-to-Text (M2T):} The model learns to translate movement into natural language descriptions, effectively performing motion captioning. All motion tokens are reserved while text tokens are randomly masked out.

\item \textbf{Motion-to-Motion (M2M):} In this setting, the model focuses solely on motion by recovering masked motion tokens, ignoring any text input. This self-supervised objective enhances the model's motion prediction and completion capabilities and enables effective classifier-free guidance for motion sequence inference.

\end{itemize}

% \paragraph{Masking Strategy.}
% We adopt a linear masking schedule: \textbf{prior to encoding}, a variable fraction of both text and motion tokens is randomly masked. 
% The masked sequences are then fed into modality-specific encoders---text into a BERT backbone and motion into a dedicated motion encoder---after which their embeddings are fused in a shared denoising backbone. 
% The corresponding classification heads reconstruct the masked text tokens and motion tokens respectively. 
% This \emph{pre-encoding masking} aligns training with inference, preserves modality-specific inductive biases, and enables effective cross-modal fusion.

\paragraph{Masking Schedule and Training Loss}
\label{sec:loss}
We train DiMo with a multi-task masked denoising objective. For each sequence $y$ and its textual description $x$, a set of positions $\mathcal{M}$ is masked, and the model predicts the masked tokens from corrupted $\tilde{y}$.  
The overall loss is masked cross-entropy:
\begin{equation}
\mathcal{L}_{\text{task}} \;=\;
\mathbb{E}_{(x,y) \sim \mathcal{D}}
\Big[ - \sum_{t \in \mathcal{M}}
\log p_\theta(y_t \mid \tilde{y}, x) \Big].
\end{equation}

% We apply a linear masking schedule, where the masking probability increases linearly with a random scalar $u \sim \mathcal{U}(0,1)$:
% \begin{equation}
% p_{\text{mask}} = u,
% \end{equation}
% This ensures that short sequences are partially corrupted while longer sequences may be heavily masked, yielding more robust learning.  
% Linear scheduling avoids overly aggressive masking at early steps and provides smoother curriculum than random or cosine schedules.
We use a linear masking schedule by sampling a per-sample masking probability
$u \sim \mathcal{U}(0,1)$, and masking each token independently with probability
$u$. This exposes the model to a broad distribution of corruption levels within
each training batch, ranging from lightly to heavily masked sequences, and
encourages robustness across varying degrees of partial observation.

\paragraph{Confidence-Guided Progressive Inference.}
% At inference, DiMo performs \textbf{confidence-guided progressive decoding}. Confidence is computed for each token directly from the model’s probability distribution. At each iteration, the model commits the most confident predictions and continues refining the rest. This mechanism achieves a balance between efficiency and quality, allowing the model to settle first on reliable predictions and progressively resolve more uncertain regions.
\begin{figure*}[t]
    \centering
    \setlength{\abovecaptionskip}{0pt} % 控制 caption 上方间距（默认一般是 10pt 左右）
    \setlength{\belowcaptionskip}{2pt} % 控制 caption 下方间距
    \includegraphics[width=0.75\textwidth]{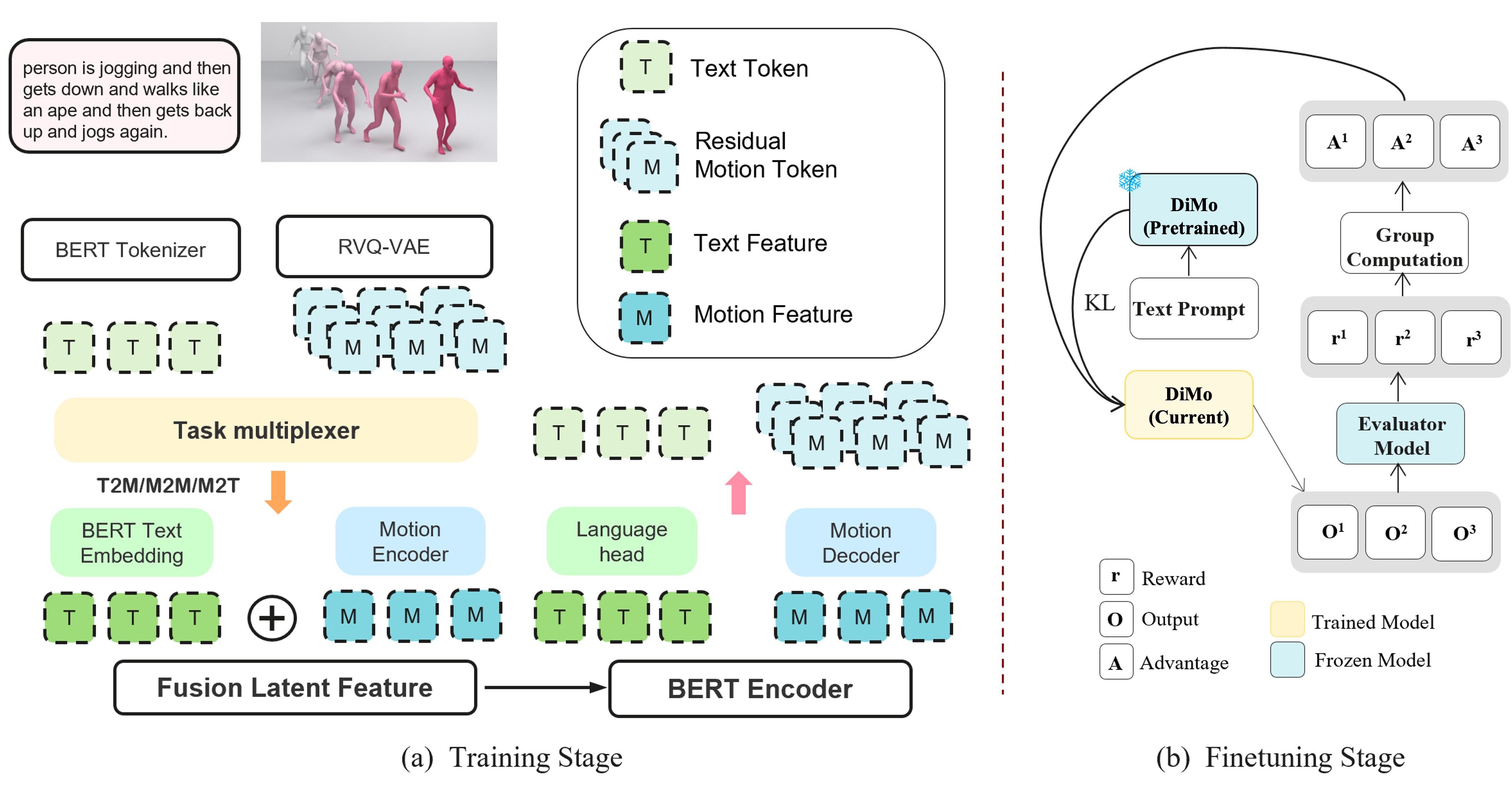}
    \caption{Overview of DiMo. Our unified framework supports text-to-motion (T2M), motion-to-text (M2T), and motion-to-motion (M2M) tasks with RVQ-based motion tokenization, multi-task masked training, confidence-guided progressive inference, and GRPO fine-tuning.}
    \vspace{-5pt}
    \label{fig:DiMo}
\end{figure*}

We adopt a confidence-guided progressive inference strategy, inspired by non-autoregressive masked decoding ~\citep{guo2023momask}. 
Given a partially masked sequence $\tilde{y}$ (motion or text tokens), the model refines it over $S$ denoising steps. 
At each step $s$, the model predicts a distribution $p_\theta(y_t \mid \tilde{y}, x)$ for all currently masked positions $t$, 
and estimates token-wise confidence by the maximum probability:
\[
c_t = \max_v p_\theta(y_t = v \mid \tilde{y}, x).
\]
Our strategy iteratively commits high-confidence tokens first and refines uncertain regions.
The full algorithm is provided in Appendix~\ref{app:progressive_inference}.

\subsection{Motion Tokenization with Residual VQ}
\label{sec:tokenization}
Following masked modeling approaches such as MoMask~\citep{guo2023momask}, 
we discretize continuous 3D motion sequences into discrete tokens via a Residual Vector Quantizer (RVQ). 
% Unlike MoMask, which employs larger codebooks, we adopt a \textbf{6-layer RVQ} with \textbf{1024 tokens per layer}, 
providing higher representational capacity while maintaining stability.  
Given a motion $M \in \mathbb{R}^{T \times J \times 3}$, the encoder produces discrete indices:
\[
z \in \{1, \ldots, N\}^{\lfloor \frac{T}{r} \rfloor \times R},
\]
where $R$ is the number of residual layers. Each layer uses an independent codebook and $N$ is the number of entities of each codebook. $r$ is the ratio of temporal compression.

\subsection{Model Architecture}
\label{sec:model}

\textbf{DiMo} is designed as a unified denoising architecture for text-to-motion (T2M), motion-to-text (M2T), and motion-to-motion (M2M) generation. The model builds upon a bidirectional masked language model while introducing modality-specific encoders and decoders for motion. This hybrid design allows DiMo to preserve the strengths of pretrained LLMs for text reasoning while explicitly modeling the structure of motion tokens. An overview of the architecture is shown in Figure~\ref{fig:DiMo}.

\paragraph{Language backbone.}
We adopt a pretrained BERT-based masked language model as the textual backbone. BERT’s bidirectional contextualization naturally aligns with our discrete diffusion-style denoising objective, in contrast to autoregressive LLMs that impose a strict left-to-right order. The backbone produces contextual embeddings for textual tokens and serves as the fusion space where text and motion features interact. We extend the tokenizer with additional special tokens to represent motion masks and padding, ensuring that the model can process multimodal sequences consistently.

\paragraph{Motion token encoder.}
Continuous motion sequences are first discretized via Residual Vector Quantization (RVQ), yielding multi-level motion tokens. To embed these tokens into the LLM’s hidden space, we introduce a dedicated motion encoder. For each RVQ codebook level, an embedding table and a lightweight Transformer encoder process the token stream, capturing hierarchical motion dynamics. The outputs are fused through learnable weights and augmented with positional embeddings. This design enables motion features to be expressed in the same representational space as text tokens, while preserving high-fidelity motion details.

\paragraph{Motion token decoder.}
Reconstruction of motion tokens is handled by a motion decoder. Each RVQ level is paired with a Transformer-based prediction head that refines backbone features and outputs logits over the motion vocabulary. By decoding tokens across RVQ levels in parallel, the decoder preserves coarse-to-fine motion fidelity while supporting efficient training and inference. An additional output projection layer maps hidden states to motion tokens, ensuring compatibility with the denoising objective.

% \paragraph{Joint modeling and denoising.}
% During training, corrupted text and motion tokens are separately embedded by the language backbone and motion encoder, concatenated into a unified sequence, and jointly processed by the BERT backbone. The fused representations are then split into text and motion branches, passed to the language head and motion decoder, respectively. Task-specific masking determines which modality is reconstructed:  
% \emph{(i)} T2M masks motion tokens while keeping text intact;  
% \emph{(ii)} M2T masks text tokens while keeping motion intact;  
% \emph{(iii)} M2M ignores text and applies self-masking to motion tokens.  
% All tasks share the same discrete diffusion-style denoising objective, enabling DiMo to unify bidirectional text-motion generation within a single architecture.

% This modular yet unified design allows DiMo to leverage pretrained language models for semantic reasoning, RVQ for high-fidelity motion representation, and diffusion-style denoising for efficient and consistent generation across modalities.

 \subsection{Reward Design}
\label{sec:reward}

To fine-tune DiMo beyond likelihood training, we design task-specific reward functions for text-to-motion (T2M) and motion-to-text (M2T), and optimize them jointly with GRPO (Sec.~\ref{sec:grpo}). The rewards are chosen to reflect semantic fidelity and modality-specific quality criteria, while remaining compatible with reinforcement learning.

\paragraph{Reward for Text-to-Motion (T2M).}
Given a text prompt $x$, the model generates a motion $\hat m$. To evaluate its semantic correctness, we utilize our pre-trained (frozen) M2T branch as a proxy evaluator to infer a pseudo caption $\hat t$ and compare it with the ground-truth caption $t$ in the CLIP embedding space:
\begin{equation}
R_{\text{T2M}}(\hat m, t) \;=\; \cos\!\big(\mathrm{CLIP}(\hat t), \,\mathrm{CLIP}(t)\big).
\end{equation}
This reward captures whether the generated motion conveys the intended textual meaning, without relying on heuristic metrics. We further explore alternative T2M reward designs, such as using external motion-to-text feature extractors. Results and ablation studies for different reward strategies are reported in Appendix~\ref{app:reward_ablation}.

\paragraph{Reward for Motion-to-Text (M2T).}
For a given motion $m$, the model outputs a caption $\hat t$. We combine verb consistency and semantic similarity, scaled by a length penalty (LP) to discourage degenerate outputs:
\vspace{-6pt}
\begin{align}
R_{\text{M2T}}(\hat t, t)
&= \lambda_{\text{verb}}\,\mathrm{VerbMatch}(\hat t, t) \nonumber\\
&\quad + \lambda_{\text{clip}}\,\cos\!\big(\mathrm{CLIP}(\hat t), \mathrm{CLIP}(t)\big)
\;\cdot\; \mathrm{LP}(|\hat t|, |t|), 
\label{eq:m2t_reward} \\[4pt]
\mathrm{LP}(|\hat t|, |t|)
&= \exp\!\Big(-\gamma \,\big| 1 - \tfrac{|\hat t|}{\max(|t|,1)} \big| \Big).
\label{eq:length_penalty}
\end{align}

Here $\mathrm{VerbMatch}$ measures verb overlap between $\hat t$ and $t$, CLIP similarity reflects semantic alignment, and $\mathrm{LP}$ penalizes captions that are disproportionately short or long. We use default weights $(\lambda_{\text{verb}}, \lambda_{\text{clip}}) = (0.3,\,0.7)$ to emphasize more on semantic consistency and $\gamma=1.0$.

% \paragraph{Joint Optimization.}
% In each training batch, every sample is first assigned to a task (e.g., T2M or M2T) according to the multi-task schedule. 
% If a sample is allocated to T2M, it is routed to the T2M reward module and evaluated by generating $G$ candidate motions, from which rewards are computed and normalized within the group before returning a loss. 
% Likewise, if a sample is allocated to M2T, it is routed to the M2T reward module, where $G$ candidate texts are generated and scored. 
% This task-aware routing ensures that each sample only contributes to its designated objective, while joint optimization over the batch allows DiMo to improve both directions in tandem.

\subsection{GRPO Objective}
\label{sec:grpo}
 
We adopt Group Relative Policy Optimization (GRPO) \citep{shao2024deepseekmath} to fine-tune DiMo. GRPO is a critic-free variant of PPO that eliminates the need for a value network by normalizing rewards within candidate groups, thereby reducing computation and memory costs.

\paragraph{Objective.}
For each input $q$ (text or motion), we sample $G$ candidate outputs $\{o_i\}_{i=1}^G$ from the old policy $\pi_{\mathrm{old}}$, 
compute rewards $r_i = R(o_i, q)$ (Sec.~\ref{sec:reward}), and update the policy $\pi_\theta$ by maximizing:
\begin{align}
J_{\text{GRPO}}(\theta)
&= \mathbb{E}_{q,\,o_i}\!\Bigg[
\frac{1}{G}\sum_{i=1}^G
\min\!\Big(
\rho_i(\theta) A_i, \nonumber\\
&\qquad\qquad
\mathrm{clip}\!\big(\rho_i(\theta),\,1-\epsilon,\,1+\epsilon\big)\,A_i
\Big)
\Bigg] \nonumber\\
&\quad - \beta\, D_{\mathrm{KL}}\!\big(
\pi_\theta(\cdot|q)\,\|\,\pi_{\mathrm{ref}}(\cdot|q)
\big).
\label{eq:grpo_main}
\end{align}

Here $\rho_i(\theta) = \tfrac{\pi_\theta(o_i \mid q)}{\pi_{\mathrm{old}}(o_i \mid q)}$ 
is the \emph{importance ratio}, i.e., the likelihood of a sampled output under the current policy relative to the old policy. 
$A_i = \tfrac{r_i - \mu}{\sigma}$ is the group-normalized advantage, 
$\epsilon=0.1$ is the clipping range, and $\beta=0.004$ is the KL coefficient. 
We found these settings to yield stable training, while larger or smaller values often led to unstable updates.

%% file: sections/5_experiments.tex
\section{Experiments}
\label{sec:experiments}

\subsection{Implementation Details}
\label{sec:setup}
% \textbf{Datasets} We evaluate DiMo on two widely used benchmarks: 
% \textbf{HumanML3D} \citep{guo2022generating} and 
% \textbf{KIT-ML} \citep{plappert2016kit}. 
% HumanML3D contains 14,616 motion sequences paired with 44,970 text descriptions, covering diverse human activities. 
% KIT-ML provides 3,911 motion clips and 6,278 text annotations. 
% Following prior works \citep{jiang2024motiongpt, guo2023momask}, HumanML3D is used as the main benchmark, and KIT-ML is adopted for additional evaluation.

% \textbf{Implementation Details.} 
Our model is trained on 16 × 32 GB Ascend 910 NPUs for 50 epochs with a mini-batch size of 32. 
Motion sequences are quantized into discrete tokens using a 6-layer RVQ tokenizer with 1024 codewords per layer.  
Text is tokenized with a pre-trained BERT-large tokenizer \citep{devlin2018bert}.
The backbone uses a hidden size of 1024 with 16 transformer layers. The downsample ratio $r$ is 4. Using a 20 fps frame rate, 1.5 tokens per frame, and a 1024-size codebook (10 bits/token), the resulting token rate is 30 tokens/s with a bitrate of 300 bits/s.

Optimization uses AdamW~\citep{DBLP:conf/iclr/LoshchilovH19} with a learning rate of $5 \times 10^{-5}$, weight decay $0.01$, and a linear warmup of 5k steps. 
Inference uses progressive unmasking with $K=20$ refinement steps for T2M and $K=30$ for M2T, classifier-free guidance (CFG) with a constant scale 3.0.  
GRPO fine-tuning is applied as a second stage (Sec.~\ref{sec:grpo}).

\subsection{Main Results}

% ---------------- HumanML3D T2M&M2T-----------------
\begin{table*}[t]

\caption{Quantitative results of Text-to-Motion and Motion-to-Text on HumanML3D.}
\vspace{-6pt}
\begin{center}
\label{tab:humanml3d_results}
\setlength{\tabcolsep}{2.8pt}
\renewcommand{\arraystretch}{1.1}
\small
\resizebox{\textwidth}{!}{%
\begin{tabular}{clccccccc|ccccccc}
\toprule
\multirow{2}{*}{\centering Category} 
& \multirow{2}{*}{\centering Method} 
& \multicolumn{7}{c|}{\textbf{T2M}} 
& \multicolumn{7}{c}{\textbf{M2T}} \\

\cmidrule(lr){3-9}  
\cmidrule(l){10-16}

&& R@1$\uparrow$ & R@2$\uparrow$ & R@3$\uparrow$ & FID$\downarrow$ & Div$\rightarrow$ & MM$\uparrow$ & MM Dist$\downarrow$
& R@1$\uparrow$ & R@3$\uparrow$ & BLEU@1$\uparrow$ & BLEU@4$\uparrow$ & ROUGE-L$\uparrow$ & CIDEr$\uparrow$ & BERTScore$\uparrow$ \\

\midrule
   & Real Motion   & 0.511 & 0.703 & 0.797 & 0.002 & 9.503 & - & 2.974 & 0.523 & 0.828 & - & - & - & - & - \\
\midrule
\multirow{10}{*}{\centering T2M Only} 
 & MDM           & -     & -     & 0.611 & 0.544 & 9.559 & 2.799 & 5.566 & - & - & - & - & - & - & - \\
 & MotionDiffuse & 0.491 & 0.681 & 0.782 & 0.630 & 9.410 & 1.553 & 3.113 & - & - & - & - & - & - & - \\
 & MLD           & 0.481 & 0.673 & 0.772 & 0.473 & 9.724 & 2.413 & 3.196 & - & - & - & - & - & - & - \\
 & MoMask        & 0.521 & 0.713 & 0.807 & 0.045 & 9.620 & 1.241 & 2.958 & - & - & - & - & - & - & - \\
 & T2M-GPT       & 0.492 & 0.679 & 0.775 & 0.141 & 9.722 & 1.831 & 3.121 & - & - & - & - & - & - & - \\
 & ReMoDiffuse   & 0.510 & 0.698 & 0.795 & 0.103 & 9.018 & 1.795 & 2.974 & - & - & - & - & - & - & - \\
 & MoGenTS       & 0.529 & 0.719 & 0.812 & 0.033 & 9.570 & -     & 2.867 & - & - & - & - & - & - & - \\
 & MotionLCM     & 0.502 & 0.698 & 0.798 & 0.304 & 9.607 & 2.259 & 3.012 & - & - & - & - & - & - & - \\
 & ReMoMask      & 0.531 & 0.722 & 0.813 & 0.099 & 9.535 & 2.823 & 2.865 & - & - & - & - & - & - & - \\
 & MaskControl   & -     &  -    & 0.805 & 0.083 & 9.395 & -     & -     & - & - & - & - & - & - & - \\

\midrule
\multirow{3}{*}{\centering \shortstack{Separated\\Model}} 
 & TM2T          & 0.424 & 0.618 & 0.729 & 1.501 & 8.589 & 2.424 & 3.467 & 0.516 & 0.823 & 48.9 & 7.0 & 38.1 & 16.8 & 32.2 \\    
 & LaMP          & 0.557 & 0.751 & 0.843 & 0.032 & 9.571 & - & 2.759 & 0.547 & 0.831 & 47.8 & 13.0 & 37.1 & 28.9 & - \\
 & MG-MotionLLM  & 0.516 & 0.706 & 0.802 & 0.303 & 9.960 & 2.125 & 2.952 & 0.592 & 0.866 & - & 8.1 & - & - & 36.7\\   

\midrule
% \cmidrule(l){2-16}
\multirow[b]{4}{*}{\centering \shortstack{Unified\\Model}} 
 & MotionGPT     & 0.492 & 0.681 & 0.778 & 0.232 & \textbf{9.528} & 2.008 & 3.096 & 0.543 & 0.827 & 48.2 & 12.5 & 37.4 & 29.2 & 32.4 \\
 & MotionGPT2    & 0.427 & 0.627 & 0.764 & 0.614  & 11.256 & 2.357 & 3.164 & 0.558 & 0.838 & 48.7 & 13.8 & 37.6 & 29.8 & 32.6 \\
 & MotionGPT3    & \textbf{0.553} & \textbf{0.747} & \textbf{0.837} & 0.208 & 9.700 & 1.018 & \textbf{2.725} & 0.573 & 0.864 & 59.1 & 19.4 & 46.2 & 28.7 & 35.2 \\
 & MoTe          & 0.548 & 0.737 & 0.825 & 0.075 & - & \textbf{2.399} & 2.867 & \textbf{0.577} & \textbf{0.871} & 46.7 & 11.2 & 37.4 & 31.5 & 30.3 \\
% \midrule
\cmidrule(l){2-16}
% \multirow{2}{*}{\centering Model}
 & \makecell[l]{Ours w/o GRPO} & 0.528 & 0.723 & 0.818 & 0.050 & 9.515 & 2.016 & 2.867 & 0.569 & 0.850 & 63.9 & 22.6 & 47.0 & 57.2 & 37.5 \\
 & \makecell[l]{Ours w/ GRPO}& 0.528 & 0.724 & 0.818 & \textbf{0.047} & 9.419  & 2.000 & 2.862 & \textbf{0.577} & 0.855 & \textbf{64.2} & \textbf{22.7} &  \textbf{47.1}  & \textbf{58.1} & \textbf{37.7} \\ 
\bottomrule
\end{tabular}%
}
\end{center}
\end{table*}

% ---------------- KIT-ML T2M&M2T-----------------
\begin{table*}[t]
\centering
\caption{Quantitative results of Text-to-Motion and Motion-to-Text on KIT-ML.}
\label{tab:kitml_results}
\setlength{\tabcolsep}{2.8pt}
\renewcommand{\arraystretch}{1.1}
\small
\resizebox{0.9\textwidth}{!}{%
\begin{tabular}{lccccccc|ccccccc}
\toprule
& \multicolumn{7}{c|}{\textbf{T2M}} & \multicolumn{7}{c}{\textbf{M2T}} \\ % <-- 加上|和&
\cmidrule(lr){2-8}  \cmidrule(l){9-15}
Method
 & R@1$\uparrow$ & R@2$\uparrow$ & R@3$\uparrow$ & FID$\downarrow$ & Div$\rightarrow$ & MM$\uparrow$ & MM Dist$\downarrow$ & R@1$\uparrow$ & R@3$\uparrow$ & BLEU@1$\uparrow$ & BLEU@4$\uparrow$ & ROUGE-L$\uparrow$ & CIDEr$\uparrow$ & BERTScore$\uparrow$ \\
\midrule
Real motion & 0.424 & 0.649 & 0.779 & 0.031 & 11.08 & - & 2.788 & 0.399 & 0.793 & - & - & - & - & - \\
\midrule
% TM2T          & 0.280 & 0.463 & 0.587 & 3.599 & 9.473 & \textbf{3.292} & 4.591 & 0.359 & 0.668 & 46.7 & \textbf{18.4} & 44.2 & \textbf{79.5} & 23.0 \\
MotionGPT     & 0.366 & 0.558 & 0.680 & 0.510 & 10.35 & 2.328 & 3.527 & - & - & - & - & - & - & - \\
MotionGPT2    & \textbf{0.427} & \textbf{0.627} & \textbf{0.764} & 0.614  & 11.256 & 2.357 & 3.164 & - & - & - & - & - & - & - \\
% MoMask        & 0.521 & 0.713 & 0.807 & 0.045 & 9.620 & 1.241 & 2.958 & - & - & - & - & - & - & - &\\
MoTe          & 0.419 & \textbf{0.627} & 0.741 & 0.256 & - & 2.615 & 3.216 & \textbf{0.421} & \textbf{0.765} & 44.9 & 14.5 & 41.8 & 55.6 & 35.9 \\
\midrule
\makecell[l]{Ours} & 0.406 & 0.620 & 0.741 & \textbf{0.206} & \textbf{10.892} & 1.690 & \textbf{2.983} & 0.396 & 0.723 & \textbf{52.5} & 17.8 & \textbf{48.0} & 68.7 & \textbf{37.7} \\
% \makecell[l]{Ours with GRPO}& - & - & - & - & -  & - & - & - & - & - & - &  -  & - & - \\ 
\bottomrule
\end{tabular}%
}
\end{table*}

\label{sec:results}
% Tables~\ref{tab:humanml3d_results} and \ref{tab:kitml_results} present results.  
% On HumanML3D, DiMo achieves the best scores across FID, R-Precision, BLEU, ROUGE-L, CIDEr, and BERTScore.  
% Table~\ref{tab:humanml3d_results} reports quantitative comparisons on the HumanML3D~\citep{guo2022generating} benchmark. Overall, DiMo (w/o GRPO) achieves strong advantages in both generation quality and text-generation metrics: for T2M quality, DiMo attains a best-in-table FID of 0.050 (before we apply GRPO finetuning), and it also shows excellent semantic consistency and diversity-related scores. For M2T, DiMo leads by a large margin on text metrics (BLEU@1 = 63.9, BLEU@4 = 22.6, ROUGE-L = 47.0, CIDEr = 57.2, BERTScore = 37.6), indicating the model not only generates high-fidelity motions but also recovers their semantic descriptions accurately. With respect to retrieval, DiMo obtains R@1 = 0.569 and R@3 = 0.850, which are competitive with the strongest baselines (some baselines retain slight advantages on individual retrieval measures). In summary, DiMo delivers a very competitive trade-off across T2M (geometric/perceptual quality) and M2T (textual quality/semantic alignment) objectives. Without GRPO, DiMo already surpasses baselines; GRPO further improves both motion realism and caption quality.
Table~\ref{tab:humanml3d_results} reports quantitative results on the
HumanML3D~\citep{guo2022generating} benchmark.
Overall, DiMo demonstrates a strong balance between motion generation
and motion understanding within a unified framework.
For Text-to-Motion (T2M), DiMo achieves one of the lowest FID scores among unified
models (0.050 without GRPO and 0.047 after GRPO), indicating that the generated
motions closely match the real-data distribution in terms of geometric and
perceptual fidelity.

On retrieval-based metrics (R@1/R@3), DiMo remains competitive but does not
consistently outperform the strongest baselines such as MotionGPT-3 or MoTe.
We observe that these metrics are highly dependent on the behavior of the
pre-trained text--motion evaluator.
Notably, even real motions only achieve 51.1\% R@1 under the same evaluator,
suggesting that higher retrieval scores may favor in-distribution motion
patterns rather than faithfully capturing semantic alignment.
This indicates that retrieval-based metrics are influenced by distributional
similarity to the evaluator’s training data and may not always reflect human
perception. To further support this observation, we include a blind human preference study in Appendix~\ref{app:user_study}.

For Motion-to-Text (M2T), where evaluation is conducted on ground-truth motions,
DiMo shows more advantages across text metrics including BLEU, ROUGE-L,
CIDEr, and BERTScore. This demonstrates that the learned motion representations preserve rich and
structured semantic information, which can be consistently mapped back to
natural language.
Overall, DiMo emphasizes perceptual realism and distributional fidelity in
motion generation, as reflected by strong FID scores, while exhibiting different
sensitivity on retrieval-based metrics that rely on a fixed evaluator.

We future validate DiMo pipeline with KIT-ML and Motion-X datasets, refer to Table~\ref{tab:kitml_results} for KIT-ML dataset. The detailed KIT-ML reports and results for Motion-X can be found at Appendix~\ref{app:motionx}. 

\begin{table}[t]
\centering
\caption{Ablation study: backbone scaling and scaling law.}
\label{tab:ablation_scalinglaw}
\setlength{\tabcolsep}{2.8pt}
\renewcommand{\arraystretch}{1.1}
\small
\resizebox{\columnwidth}{!}{%
\begin{tabular}{lccc|ccccc}
\toprule
& \multicolumn{3}{c|}{Text-to-Motion} & \multicolumn{5}{c}{Motion-to-Text} \\
\cmidrule(lr){2-4} \cmidrule(l){5-9}
Backbone
 & R@1$\uparrow$ & R@3$\uparrow$ & FID$\downarrow$ & R@1$\uparrow$ & R@3$\uparrow$ & BLEU@1$\uparrow$ & ROUGE-L$\uparrow$ & CIDEr$\uparrow$ \\
\midrule
ALBERT      & 0.455  & 0.753 & 0.947 & 0.528 & 0.821 & 63.1 & 46.1 & 54.5 \\
BERT-base   & 0.493  & 0.780 & 0.135 & \textbf{0.582} & \textbf{0.857} & \textbf{65.0} & \textbf{47.6} & \textbf{58.9} \\
BERT-large  & \textbf{0.528} & \textbf{0.818} & \textbf{0.050} & 0.569 & 0.850 & 63.9 & 47.0 & 57.2 \\
\bottomrule
\end{tabular}%
}
\vspace{-15 pt}
\end{table}

% ---------------- Ablation computation vs quality-----------------
\begin{table*}[t]
\centering
% \caption{Ablation on denoising steps for BERT-large backbone (\texttt{motion\_infer\_times}, \texttt{text\_infer\_times}). BLEU@1, BLEU@4, ROUGE, CIDEr, BERTScore in percentage.}
\caption{Ablation on computation vs quality. Text metrics (BLEU@1, BLEU@4, ROUGE-L, CIDEr, BERTScore) are in percentage. Our model is evaluated under different denoising steps with BERT-large backbone.}
\label{tab:denoising_steps}
\setlength{\tabcolsep}{2.8pt}
\renewcommand{\arraystretch}{1.1}
\small
\resizebox{\textwidth}{!}{%
\begin{tabular}{cccccc|ccccccc|cc}
\toprule
& \multicolumn{5}{c|}{Text-to-Motion} & \multicolumn{7}{c}{Motion-to-Text} & \multicolumn{2}{c}{Computational Cost}\\
\cmidrule(lr){2-6} \cmidrule(l){7-13} \cmidrule(l){14-15}
Method & R@1$\uparrow$ & R@2$\uparrow$ & R@3$\uparrow$ & FID$\downarrow$ & Latency(s) & R@1$\uparrow$ & R@3$\uparrow$ & BLEU@1$\uparrow$ & BLEU@4$\uparrow$ & ROUGE$\uparrow$ & BERTScore$\uparrow$ & Latency(s) & \string#Params & FLOPs/sample \\
\midrule
MotionGPT     & 0.492 & 0.681 & 0.778 & 0.232 & 1.04 & 0.543 & 0.827 & 48.2 & 12.5 & 37.4 & 32.4 & 0.48 & 220M & 7.45T\\
MotionGPT-3   & \textbf{0.553} & \textbf{0.747} & \textbf{0.837} & 0.208 & 1.02 & 0.573 & 0.864 & 59.1 & 19.4 & 46.2 & 35.2 & 1.30 & 238M & 11T  \\
MG-MotionLLM  & 0.516 & 0.706 & 0.802 & 0.303 & 1.09 & \textbf{0.592} & \textbf{0.866} &   -  &  8.1 &   -  & 36.7 & 0.60 & 220M & 1.66T\\
\midrule
Ours(5 steps) & 0.523 & 0.719 & 0.812 & 0.120 & 0.41 & 0.465 & 0.744 & 54.7 & 16.2 & 42.2 & 19.2 & 0.21 & 473M & 0.64T\\
Ours(10 steps)& 0.527 & 0.724 & 0.817 & 0.068 & 0.76 & 0.510 & 0.795 & 56.8 & 19.1 & 44.9 & 26.4 & 0.38 & 473M & 1.28T\\
Ours(20 steps)& 0.528 & 0.723 & 0.818 & \textbf{0.050} & 1.55 & 0.568 & 0.845 & 62.5 & 22.0 & \textbf{47.3} & 35.4 & 0.81 & 473M & 2.56T\\
Ours(30 steps)& 0.524 & 0.720 & 0.815 & 0.052 & 2.39 & 0.569 & 0.850 & \textbf{63.9} & \textbf{22.6} & 47.0 & \textbf{37.5} & 1.18 & 473M & 3.84T\\
% 40  & 0.524 & 0.720 & 0.815 & 0.053 & xxxx & 0.584 & 0.858 & 63.9 & 22.5 & 46.6  & 39.0 & xxxx & xxxx & xxxx\\

\bottomrule
\end{tabular}%
}
\end{table*}

% ---------------- Ablation RVQ layer-----------------
\begin{table*}[t]
\centering
\caption{Ablation study: model performance under different RVQ layer.}
\label{tab:ablation_rvq}
\setlength{\tabcolsep}{2.8pt}
\renewcommand{\arraystretch}{1.1}
\small
\resizebox{0.87\textwidth}{!}{%
\begin{tabular}{cc|cccc|ccccccc}
\toprule
\multirow{2}{*}{\shortstack{\#RVQ\\Layer}} 
& \multicolumn{1}{c|}{Reconstruction}& \multicolumn{4}{c|}{Text-to-Motion} & \multicolumn{7}{c}{Motion-to-Text} \\
\cmidrule(l){2-2}\cmidrule(l){3-6} \cmidrule(l){7-13}

& MSE$\downarrow$ & R@1$\uparrow$ & R@2$\uparrow$ & R@3$\uparrow$ & FID$\downarrow$ & R@1$\uparrow$ & R@3$\uparrow$ & BLEU@1$\uparrow$ & BLEU@4$\uparrow$ & ROUGE-L$\uparrow$ & CIDEr$\uparrow$ & BERTScore$\uparrow$ \\
\midrule
1 & 0.0830 & 0.515 & 0.711 & 0.806 & 0.200 & 0.558 & 0.841 & 62.8 & 22.0 & 46.5 & 55.0 & 36.2 \\
4 & 0.0215 & 0.527 & 0.722 & 0.821 & 0.076 & \textbf{0.579} & 0.845 & \textbf{64.2} & \textbf{22.8} & \textbf{47.2} & \textbf{58.3} & 37.0 \\
6 & 0.0118 & 0.528 & 0.723 & 0.818 & \textbf{0.050} & 0.569 & \textbf{0.850} & 63.9 & 22.6 & 47.0 & 57.2 & \textbf{37.5} \\
8 & \textbf{0.0101} & \textbf{0.531} & \textbf{0.729} & \textbf{0.824} & 0.121 & 0.573 & 0.847 & 63.8 & 22.3 & 47.0 & 57.4 & 37.1 \\
\bottomrule
\end{tabular}%
}
\end{table*}

\subsection{Ablation Studies}
\label{sec:ablation}

\textbf{Ablation — Backbone scaling and scaling law.} Table \ref{tab:ablation_scalinglaw} compares three masked-language backbones (ALBERT, BERT-base, BERT-large) to study the correlation between model size and generation quality. We observe a monotonic improvement on T2M task with backbone scale (BERT-large $>$ BERT-base $>$ ALBERT), yielding consistent gains across key metrics such as R-Precision, FID. However, in terms of text performance, bert-base shows better results, 
likely because the textual corpus in HumanML3D is relatively limited. 
As future work, it is necessary to investigate the scaling laws on larger-scale motion generation datasets.

\textbf{Latency–quality Trade-off.} The progressive denoising mechanism of discrete diffusion enables a smooth trade-off between inference latency and generation quality by tuning the number of sampling steps and confidence thresholds. We measure latency and core quality metrics (FID, BLEU, R@1, and human preference) under several sampling budgets (e.g., $K=30,20,10$; see Table~\ref{tab:denoising_steps}). Results indicate that reducing $K$ from 30 to 5 substantially cuts inference time while only incurring modest quality degradation. Despite having a larger model size than prior baselines, our model achieves a better quality–latency trade-off: very few steps (e.g., $K=5$) already outperform existing methods in T2M FID with minimal latency, while more steps (10–30) further improve quality metrics at the cost of higher but still reasonable latency. This demonstrates that the number of refinement (denoising) steps provides flexible operating points balancing speed and quality beyond previous approaches.

\textbf{Effect of RVQ Stage Depth.} Table~\ref{tab:ablation_rvq} shows that multi-stage residual quantization significantly improves reconstruction fidelity and downstream generation/understanding performance. A 6-stage RVQ consistently outperforms single-stage quantization on FID, R-Precision, and all text-based metrics. The hierarchical residual codebooks allow the first stage to capture coarse, low-frequency body structure while later stages encode high-frequency details. This frequency-wise decomposition substantially reduces quantization error for fine details and provides the discrete diffusion backbone with more informative tokens, yielding motions that are clearer in both semantics and detail and producing more accurate textual descriptions. Reconstruction MSE decreases monotonically with increasing RVQ depth, confirming that deeper quantization better preserves motion details. Downstream T2M/M2T metrics improve up to 4–6 layers but saturate or fluctuate slightly beyond that, likely due to the trade-off between tighter reconstruction and more complex codebooks. Here we use refinement steps $K=20$ for T2M generation and $K=30$ for M2T generation. 

Besides the above mentioned experiments, we also thoroughly ablate different masking schedules across modalities, different task proportions, scales of classifier-free guidance for motion sequence generation, and study the effectiveness of multi-task training. More discussion details can be found in Appendix~\ref{sec:ablations}.

\begin{figure*}[t]
    \centering
    \setlength{\abovecaptionskip}{6pt} % 控制 caption 上方间距（默认一般是 10pt 左右）
    \setlength{\belowcaptionskip}{4pt} % 控制 caption 下方间距
    \includegraphics[width=0.9\textwidth]{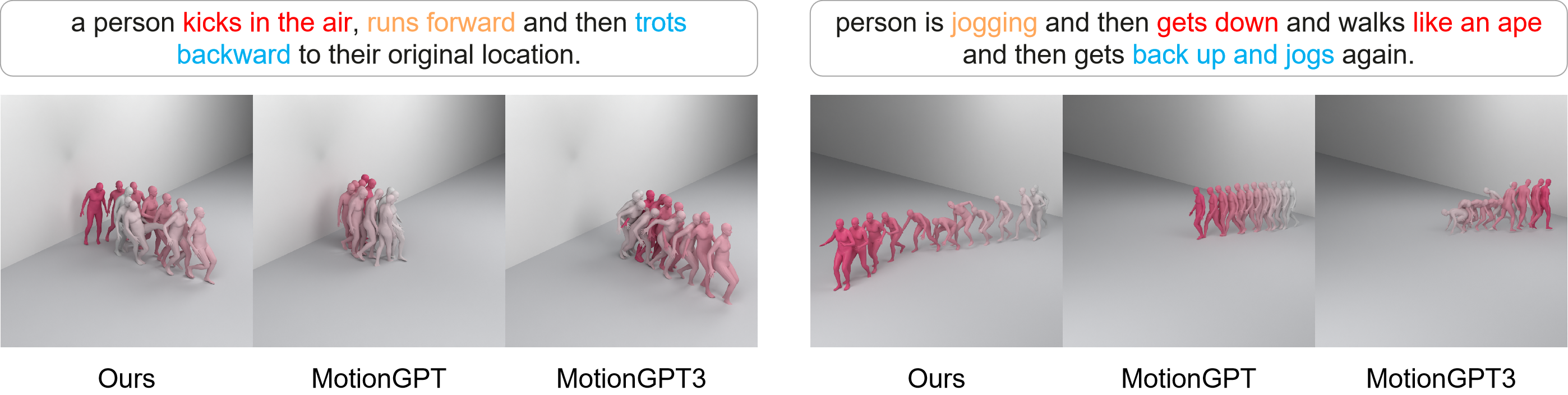}
    \caption{Text-to-motion comparison: DiMo generates more coherent and semantically aligned motions.}
        % \vspace{-6pt}
    \label{fig:t2m}
\end{figure*}

% \begin{figure}
%     \centering
%     \setlength{\abovecaptionskip}{4pt} % 控制 caption 上方间距（默认一般是 10pt 左右）
%     \setlength{\belowcaptionskip}{0pt} % 控制 caption 下方间距
%     \includegraphics[width=0.9\columnwidth]{Assets/images/Picture3.jpg}
%     \caption{Text-to-motion comparison: DiMo generates more coherent and semantically aligned motions.}
%      \vspace{-10pt}
%     \label{fig:t2m}
% \end{figure}

\begin{figure*}[t]
    \centering
    \setlength{\abovecaptionskip}{6pt} % 控制 caption 上方间距（默认一般是 10pt 左右）
    \setlength{\belowcaptionskip}{0pt} % 控制 caption 下方间距
    \includegraphics[width=0.9\textwidth]{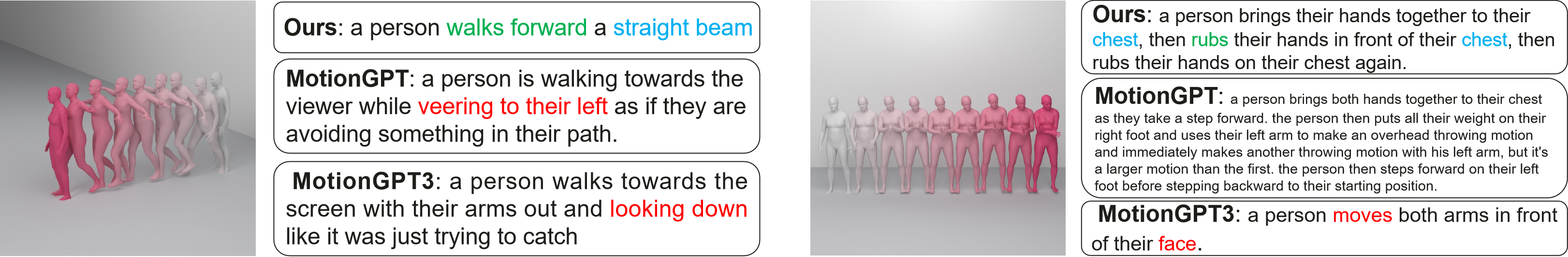}
    % \vspace{-6pt}
    \caption{Motion-to-text comparison: DiMo produces concise and accurate action descriptions.}
    \label{fig:m2t}
\end{figure*}

\subsection{Qualitative Results}   
\label{sec:qualitative}

% \begin{figure}
%     \centering
%     \setlength{\abovecaptionskip}{4pt} % 控制 caption 上方间距（默认一般是 10pt 左右）
%     \setlength{\belowcaptionskip}{0pt} % 控制 caption 下方间距
%     \includegraphics[width=0.9\columnwidth]{Assets/images/Picture1.jpg}
%     % \vspace{-6pt}
%     \caption{Motion-to-text comparison: DiMo produces concise and accurate action descriptions.}
%     \vspace{-10pt}
%     \label{fig:m2t}
% \end{figure}

We present qualitative examples to further illustrate the characteristics of DiMo’s bidirectional denoising paradigm. 
Unlike one-pass autoregressive generation, DiMo can iteratively refine outputs by re-masking and re-denoising subsequences. 
This flexibility enables progressive improvement of motion quality and the ability to adjust local segments without regenerating the entire sequence. 

Figure~\ref{fig:t2m} shows a text-to-motion example: DiMo generates motions that follow the caption and preserve coherent body dynamics.
Figure~\ref{fig:m2t} shows a motion-to-text case: given a motion sequence, DiMo produces concise captions that capture key actions and their temporal order.

These results indicate that DiMo handles both directions within a unified framework.
Compared with autoregressive approaches such as MotionGPT and MotionGPT3, DiMo tends to preserve fine-grained motion details throughout the sequence, maintaining later-stage semantics without relying on strong early-frame cues. We have porvided  more qualitative comparisons with a structured demo page in the supplementary material.

%% file: sections/6_applications.tex
\section{Applications}
\label{sec:applications}
% \subsection{Motion Completion}
% \label{sec:app_completion}

% \textbf{Motion Inbetweening/Completion}. Given the beginning and ending segments of a motion sequence, DiMo reconstructs the masked middle portion to ensure temporal coherence and kinematic naturalness. Unlike text-conditioned methods, it operates in a text-free manner by iteratively denoising the missing region. As shown in Figure\ref{fig:application}(a), this process yields smooth transitions and semantically consistent motions.

% \textbf{Motion Continuation/Prediction}. DiMo supports motion continuation, where a motion and its textual description are given and additional semantic content is appended to the text to extend the sequence. The model uses the new linguistic cues to generate successive motion segments, continuing the motion naturally. Figure~\ref{fig:application}(b) motion continuation demonstrates coherent extensions.

% \textbf{Caption Correction}. Beyond motion inbetweening and continuation, DiMo also supports motion-guided caption correction. When a caption mismatches the underlying action (e.g., describing walking as running), the model refines the text to align with the observed motion. As shown in Figure~\ref{fig:application}(c), iterative denoising progressively enforces temporal and semantic consistency.

\textbf{Motion Inbetweening/Completion}. Given the beginning and ending segments of a motion sequence, DiMo reconstructs the masked middle portion to ensure temporal coherence and kinematic naturalness. Unlike text-conditioned methods, it operates in a text-free manner by iteratively denoising the missing region. 
This allows DiMo to flexibly handle partial observations without relying on explicit linguistic supervision.
As shown in Figure~\ref{fig:application}(a), this process yields smooth transitions and semantically consistent motions.

\textbf{Motion Continuation/Prediction}. DiMo supports motion continuation, where a motion and its textual description are given and additional semantic content is appended to the text to extend the sequence. The model uses the new linguistic cues to generate successive motion segments, continuing the motion naturally. 
This setting highlights DiMo’s ability to fuse evolving language semantics with long-horizon motion dynamics.
Figure~\ref{fig:application}(b) demonstrates coherent motion extensions.

\textbf{Caption Correction}. Beyond motion inbetweening and continuation, DiMo also supports motion-guided caption correction. When a caption mismatches the underlying action (e.g., describing walking as running), the model refines the text to align with the observed motion. 
By iteratively denoising text tokens conditioned on motion, DiMo enforces tighter cross-modal consistency.
As shown in Figure~\ref{fig:application}(c), this process progressively corrects semantic discrepancies.

%% file: sections/7_conclusions.tex
\begin{figure}[!t]
    \centering
    \setlength{\abovecaptionskip}{4pt} % 控制 caption 上方间距（默认一般是 10pt 左右）
    \setlength{\belowcaptionskip}{0pt} % 控制 caption 下方间距
    \includegraphics[width=0.87\columnwidth]{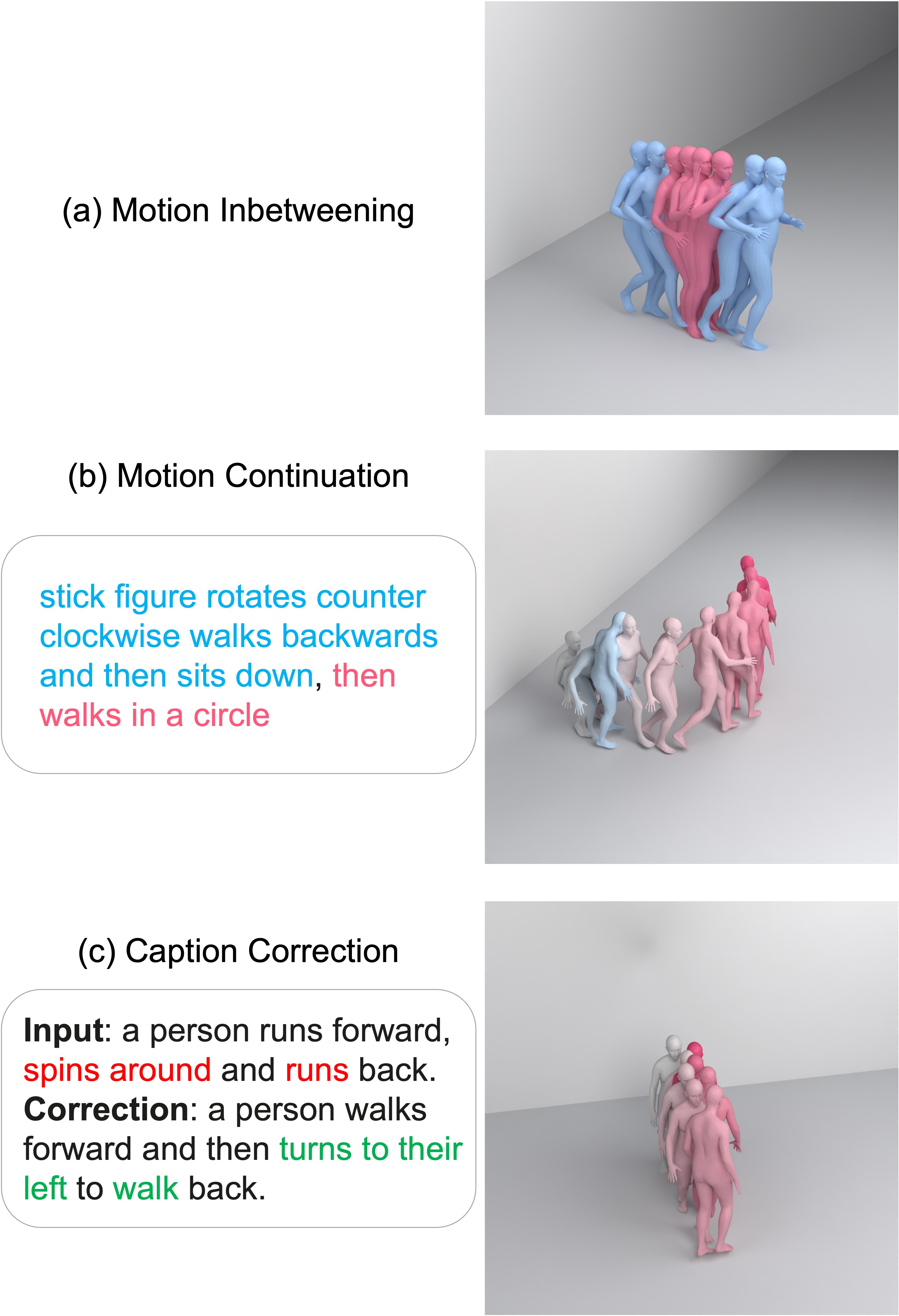}
    % \vspace{-6pt}
    \caption{Application: Examples of downstream tasks enabled by DiMo}
    \vspace{-16pt}
    \label{fig:application}
\end{figure}

\section{Conclusions}
\label{sec:conclusions}
 
% We presented \textbf{DiMo}, a unified discrete diffusion framework for
% bidirectional text-to-motion and motion-to-text modeling.
% By integrating masked progressive refinement, classifier-free guidance,
% and GRPO-based fine-tuning, DiMo supports efficient parallel decoding and
% achieves a strong balance between motion generation quality and semantic
% understanding.
% Experiments on HumanML3D and KIT-ML show that DiMo performs competitively
% with prior unified motion--language models, particularly in terms of
% perceptual motion fidelity and text generation accuracy.
% Beyond standard T2M and M2T tasks, the same framework naturally extends to
% motion completion and prediction plus caption correction without architectural changes.
% Future work includes scaling to richer motion representations (e.g., meshes
% or multi-agent scenarios) and exploring more robust external reward models
% to further improve controllability and semantic alignment.

We have presented \textbf{DiMo}, a unified discrete diffusion framework for
bidirectional text-to-motion (T2M) and motion-to-text (M2T) modeling.
By combining masked progressive refinement with classifier-free guidance,
DiMo enables efficient parallel decoding and supports motion generation and
understanding within a single framework.
With additional GRPO-based fine-tuning, the model benefits from task-specific
reinforcement signals that consistently improve semantic alignment and output
quality across both directions.
Experiments on HumanML3D and KIT-ML demonstrate that DiMo
achieves a strong and well-balanced performance compared to prior unified
motion--language models, particularly in terms of perceptual motion quality
and text generation accuracy. More discussions are at Appendix~\ref{app:limitations}. 

Beyond standard T2M and M2T tasks, DiMo naturally extends to motion completion,
prediction, and caption correction without architectural changes, highlighting the
flexibility of the proposed paradigm.
Future work includes scaling the framework beyond skeletal motion to
mesh-based or multi-agent settings, as well as exploring more robust
external reward models for GRPO to further enhance controllability,
robustness, and semantic faithfulness. 

\section*{Impact Statement}
This paper presents work whose goal is to advance the field of machine learning. There are many potential societal consequences of our work, none of which we feel must be specifically highlighted here.

%% file: sections/8_appendix.tex
\appendix
{\LARGE\sc {Appendix for DiMo}\par}

% \section*{Reproducibility Statement}
% We have made extensive efforts to ensure the reproducibility of DiMo. 
% The datasets used in our experiments (HumanML3D) are publicly available. 
% Details of our pipeline, motion encoder/decoder modules, and the BERT-based language backbone are presented in Sec.~\ref{sec:model}. 
% Training procedures, including optimization settings, masking schedules, and reinforcement fine-tuning with GRPO, are documented in Sec.~\ref{sec:grpo}. 
% Evaluation protocols and metrics for T2M, M2T, and motion completion will be discussed in details at 
% Comprehensive descriptions of model architectures and hyperparameter settings are further provided in Appendix~\ref{sec:metrics}. 

\section*{LLM Usage}
A large language model was used as a writing assistant to improve the clarity and readability of the manuscript. 
All final research decisions and methodological contributions were made and verified by the authors.

\section{Confidence-Guided Progressive Inference}
\label{app:progressive_inference}

During inference, DiMo follows a progressive unmasking strategy guided by prediction confidence.
Starting from a fully masked or partially corrupted sequence, the model iteratively fills masked positions
according to a predefined schedule $\{k_s\}_{s=1}^S$, where $k_s$ controls how many tokens are committed at step $s$.

At each iteration, the model predicts token distributions for all masked positions and computes a confidence
score as the maximum predicted probability.
A subset of masked positions with the highest confidence is then selected and committed,
while the remaining positions stay masked for subsequent iterations.

This strategy has two key advantages:
(1) high-confidence tokens are fixed early, anchoring the global structure of the sequence;
(2) uncertain regions are deferred and refined in later steps, reducing error accumulation
and improving global coherence.
After $S$ iterations, all tokens are committed, yielding the final generated sequence.

\begin{algorithm}[h] 
\caption{Confidence-Guided Progressive Inference} 
\label{alg:progressive} 
\begin{algorithmic}[1] 
\REQUIRE Corrupted sequence $\tilde{y}$, condition $x$, inference steps $S$ 
\ENSURE Generated sequence $y$ 
\FOR{$s = 1$ to $S$} 
\STATE Predict logits $p_\theta(y_t \mid \tilde{y}, x)$ for all masked $t$ 
\STATE Compute confidence $c_t = \max_v p_\theta(y_t=v \mid \tilde{y}, x)$ 
\STATE Select top-$k_s$ masked positions with highest $c_t$ 
\STATE Commit: $\tilde{y}_t \leftarrow \arg\max_v p_\theta(y_t=v \mid \tilde{y}, x)$ \ENDFOR 
\STATE \textbf{return} $y$
\end{algorithmic} 
\end{algorithm}

\section{Evaluation Metrics}
\label{sec:metrics}
For \textbf{Text-to-Motion (T2M)}, follow existing works \cite{guo2022tm2t, guo2023momask, jiang2024motiongpt, wang2024motiongpt2, zhu2025motiongpt3}, we evaluate motion quality and text–motion alignment. Motion realism is measured by Fréchet Inception Distance (FID), while R-Precision (R@1/2/3) and Multimodal Distance (MM Dist) assess semantic consistency between motion and text. We further report Diversity (Div) to capture variation across generated motions, and Multi-Modality (MM) to measure the ability to produce multiple plausible outputs for the same text. For \textbf{Motion-to-Text (M2T)}, follow existing works \cite{guo2022tm2t, guo2023momask, jiang2024motiongpt, wang2024motiongpt2, zhu2025motiongpt3}, we adopt standard captioning metrics, including BLEU@1/4, ROUGE-L, CIDEr, and BERTScore, which jointly evaluate lexical overlap and semantic similarity to reference descriptions. In addition, we report R-Precision to measure alignment between generated texts and their corresponding motions. The detailed formulations of each metric are introduced below.

\textbf{R-Precision.} 
R-Precision measures retrieval performance by computing the fraction of relevant items within the top-$R$ retrieved results. In text-to-motion, this means retrieving the correct motion from a database given a text query, or vice versa.  

\begin{equation}
R\text{-Prec} = \frac{| \text{Rel} \cap \text{Top-}R |}{R}
\end{equation}

where $\text{Rel}$ is the set of relevant items (ground-truth matches) and $\text{Top-}R$ is the set of retrieved items at rank $R$. The metric ranges from 0 to 1, with higher values indicating better retrieval accuracy.

\textbf{Fr\'echet Inception Distance (FID).} 
FID~\cite{guo2022generating} measures the distributional distance between real and generated samples in a feature space, capturing both mean and covariance statistics. Lower FID indicates that generated samples are closer to real samples in distribution.  

\begin{equation}
\text{FID} = \| \mu_r - \mu_g \|_2^2 + \mathrm{Tr}\big(\Sigma_r + \Sigma_g - 2(\Sigma_r \Sigma_g)^{1/2}\big)
\end{equation}

where $\mu_r, \Sigma_r$ are the mean and covariance of real samples in the feature space, and $\mu_g, \Sigma_g$ are the corresponding statistics of generated samples. The first term measures the distance between means, while the second term accounts for differences in covariance structure.

\textbf{Diversity.} 
Diversity~\cite{guo2022generating} quantifies the average pairwise distance between generated samples in the feature space, reflecting variability among generated motions. Higher values indicate more diverse generations.  

\begin{equation}
\text{Div} = \frac{2}{M(M-1)} \sum_{i<j} \| f(x_i) - f(x_j) \|_2
\end{equation}

where $\{x_i\}_{i=1}^M$ are generated samples and $f(\cdot)$ is a feature extractor such as a motion encoder. The summation averages the pairwise Euclidean distance across all distinct sample pairs.

\textbf{Multi-Modality.} 
Multi-Modality~\cite{guo2022generating} measures the variability of outputs generated from the same input condition, capturing the model’s ability to produce multiple plausible outputs. It is computed as the average pairwise distance between multiple samples generated for the same condition.  

\begin{equation}
\text{MM} = \frac{2}{K(K-1)} \sum_{i<j} \| f(x_i^t) - f(x_j^t) \|_2
\end{equation}

where $\{x_i^t\}_{i=1}^K$ are the generated samples conditioned on the same text $t$, and $f(\cdot)$ is a feature extractor. The metric encourages both relevance to the condition and diversity among outputs.

\textbf{Multimodal Distance.} 
Multimodal Distance~\cite{guo2022generating} evaluates how closely aligned text and motion representations are in a shared embedding space. Lower values indicate better alignment between modalities.  

\begin{equation}
\text{MM Dist} = \frac{1}{N} \sum_{i=1}^N d(f_{\text{text}}(t_i), f_{\text{motion}}(m_i))
\end{equation}

where $\{(t_i, m_i)\}_{i=1}^N$ are paired text and motion samples, $f_{\text{text}}$ and $f_{\text{motion}}$ are embedding functions for text and motion respectively, and $d(\cdot, \cdot)$ is a distance function such as Euclidean distance or cosine distance.

\textbf{BLEU.} 
BLEU~\cite{papineni2002bleu} evaluates the n\text{-}gram precision of generated text against reference text, penalized by a brevity term to avoid favoring short outputs. 
The most common setup is BLEU@4, which considers up to 4\text{-}gram matches.  

\begin{equation}
\text{BLEU@N} = \text{BP} \cdot \exp\left(\frac{1}{N} \sum_{n=1}^{N} \log p_{n}\right)
\end{equation}

where $N$ is the maximum n\text{-}gram order (commonly $N=4$), and $p_n$ is the modified n\text{-}gram precision defined as  

\begin{equation}
p_{n} = 
\frac{\sum_{\text{ngram} \in C} \min \Big(\text{Count}_{C}(\text{ngram}), \max_{R \in \text{Refs}} \text{Count}_{R}(\text{ngram}) \Big)}
{\sum_{\text{ngram} \in C} \text{Count}_{C}(\text{ngram})}.
\end{equation}

The brevity penalty (BP) is applied to discourage very short candidates:
\begin{equation}
\text{BP} =
\begin{cases}
1 & \text{if } c > r, \\
e^{(1-r/c)} & \text{if } c \leq r,
\end{cases}
\end{equation}

where $C$ denotes the candidate sentence, $\text{Refs}$ the set of reference sentences, $\text{Count}_{C}(\text{ngram})$ the number of times an n\text{-}gram appears in the candidate, and $\text{Count}_{R}(\text{ngram})$ the count of the same n\text{-}gram in reference $R$. The clipping operation $\min(\cdot)$ ensures that the n\text{-}gram precision does not reward repeated n\text{-}grams beyond what occurs in references. Here $c$ is the length (in tokens) of the candidate sentence, and $r$ is the effective reference length, chosen as the reference length closest to $c$.

\textbf{ROUGE-L.} 
ROUGE-L~\cite{lin2004rouge} measures the quality of generated text by computing the longest common subsequence (LCS) between candidate and reference, which captures sentence-level fluency without requiring consecutive matches. The score combines precision and recall of the LCS using an $F$-measure formulation.  

\begin{equation}
\text{ROUGE-L} = \frac{(1+\beta^2) \cdot P_{LCS} \cdot R_{LCS}}{R_{LCS} + \beta^2 \cdot P_{LCS}}
\end{equation}

where $R_{LCS} = \frac{LCS(c, r)}{|r|}$ and $P_{LCS} = \frac{LCS(c, r)}{|c|}$, with $LCS(c, r)$ denoting the length of the longest common subsequence between candidate $c$ and reference $r$. The parameter $\beta$ controls the relative importance of recall and precision (commonly $\beta=1$ for equal weighting). Here $|c|$ is the candidate length in tokens and $|r|$ is the reference length in tokens.

\textbf{CIDEr.} 
CIDEr~\cite{vedantam2015cider} evaluates the similarity of a candidate sentence against multiple references using a TF-IDF weighted n\text{-}gram representation. It is designed to capture consensus among reference captions and reduce the effect of common n\text{-}grams.  

\begin{equation}
\text{CIDEr}(c, S) = \frac{1}{|S|} \sum_{s \in S} \frac{g(c) \cdot g(s)}{\|g(c)\| \|g(s)\|}
\end{equation}

where $c$ is the candidate sentence, $S$ is the set of reference sentences, and $g(x)$ denotes the TF-IDF vector representation of n\text{-}grams extracted from text $x$. The numerator $g(c) \cdot g(s)$ is the dot product between the candidate and reference vectors, while the denominator normalizes by their Euclidean norms to compute cosine similarity.  

\textbf{BERTScore.} 
BERTScore~\cite{zhang2019bertscore} evaluates the semantic similarity between generated and reference sentences using contextual embeddings from a pretrained language model such as BERT. It computes the average maximum similarity between tokens across candidate and reference.  

\begin{equation}
\text{BERTScore}(c, r) = \frac{1}{|c|} \sum_{x \in c} \max_{y \in r} \cos\big(f(x), f(y)\big)
\end{equation}

where $c$ is the candidate sentence, $r$ is the reference sentence, $f(\cdot)$ is the contextual embedding function from BERT, and $\cos(\cdot, \cdot)$ denotes cosine similarity. For each token $x$ in the candidate, the most similar token $y$ in the reference is found in embedding space, and the similarities are averaged across all candidate tokens.  

\section{Additional Analysis and User Studies}
\label{app:user_study}
\begin{table}[t]
\centering
\caption{Text-to-Motion quantitative results on HumanML3D.}
\label{tab:t2m_main_appendix}
\small
\begin{tabular}{lccccccc}
\toprule
Method & R@1$\uparrow$ & R@2$\uparrow$ & R@3$\uparrow$ & FID$\downarrow$ & Div$\rightarrow$ & MM$\uparrow$ & MM Dist$\downarrow$ \\
\midrule
Real motions     & 0.511 & 0.703 & 0.797 & 0.002 & 9.503 & --    & 2.974 \\
TM2T             & 0.424 & 0.618 & 0.729 & 1.501 & 8.589 & 2.424 & 3.467 \\
MotionGPT        & 0.492 & 0.681 & 0.778 & 0.232 & \textbf{9.528} & 2.008 & 3.096 \\
MotionGPT2       & 0.427 & 0.627 & 0.764 & 0.614 & 11.256 & 2.357 & 3.164 \\
MotionGPT3       & \textbf{0.553} & \textbf{0.747} & \textbf{0.837} & 0.208 & 9.700 & 1.018 & \textbf{2.725} \\
MG-MotionLLM     & 0.516 & 0.706 & 0.802 & 0.303 & 9.960 & 2.125 & 2.952 \\
MoTe             & 0.548 & 0.737 & 0.825 & 0.075 & -- & 2.399 & 2.867 \\
Ours w/o GRPO    & 0.528 & 0.723 & 0.818 & 0.050 & 9.515 & 2.016 & 2.867 \\
Ours w/ GRPO     & 0.528 & 0.724 & 0.818 & \textbf{0.047} & 9.419 & 2.000 & 2.862 \\
\bottomrule
\end{tabular}
\end{table}

In the Text-to-Motion literature, Fr\'echet Inception Distance (FID) is used for evaluating motion generation quality, as it directly measures the distributional similarity between generated motions and real data.
On HumanML3D, DiMo achieves a clear improvement in FID over prior unified frameworks (e.g., 0.047 vs. 0.075 for MoTe), indicating that its generated motions are closer to the real motion distribution. In contrast, R-Precision depend heavily on the quality and generalization ability of the text–motion evaluator used for scoring. To investigate how well the r-precision evaluation mechanism aligns with human choices, we conduct the user study with the following setup: 1) Each trial showed the text prompt and [two/three] rendered motion animations side-by-side. No method names were shown.2) Randomization and blinding. The left/right (or order) assignment was randomized per trial.
Raters were blinded to the source model.

As shown in Table~\ref{tab:t2m_main_appendix}, Real motions only achieve an R@1 of 51.1\%, which is lower than MotionGPT-3, MG-MotionLLM, MoTe, and our method under the same evaluator. This does not mean that these models generate motions more faithful to the text prompts than the ground truth. Instead, it indicates that the evaluator has limited generalization: it more easily recognizes in-domain motions similar to those seen during training, while its retrieval performance deteriorates for motions that it has not seen before. Consequently, a higher R-Precision may largely reflect that the generated sequences resemble training motions, rather than guaranteeing better semantic alignment with the text. To further validate this hypothesis, we conduct human preference studies (Table~\ref{tab:user_study}):1)Ground Truth vs. MotionGPT-3 (200 test samples, given same text description ), 2) Ground Truth vs. Ours (200 test samples), and 3) MotionGPT vs. MotionGPT-3 vs. Ours (1060 test samples); given same text description, let user pick which motion is more aligned with text.

While the numerical R-Precision scores suggest the ordering MotionGPT-3 \textgreater Ours \textgreater Ground Truth, human raters consistently prefer MotionGPT-3 \textless Ours \textless Ground Truth. This discrepancy highlights the unreliability of R-Precision as an absolute quality measure among strong methods. Therefore, on the T2M task, considering both FID and human preferences, our approach delivers clearly improved motion quality and text–motion consistency compared to existing work.

\begin{table}[t]
\centering
\caption{Human preference study results on Text-to-Motion generation.}
\label{tab:user_study}
\small
\begin{tabular}{l|c|c|c|c}
\toprule
Setting & \#Samples & Option 1 & Option 2 & Option 3 \\
\midrule
GT vs. MotionGPT-3 & 200 & GT: 74\% & MotionGPT-3: 26\% & - \\
GT vs. Ours        & 200 & GT: 60\% & Ours: 40\% & - \\
MotionGPT vs. MotionGPT-3 vs. Ours & 1060 & MotionGPT: 27.6\% & MotionGPT-3: 32.8\% & Ours: 39.6\% \\
\bottomrule
\end{tabular}
\end{table}

\section{More Quantitative Results}
\label{sec:ablations}
 
\subsection{Ablation on T2M Reward Design}
\label{app:reward_ablation}
\label{app:reward_ablation}
To study how reward design affects GRPO fine-tuning for Text-to-Motion (T2M),
we compare three variants under the same backbone and training protocol:
(i) \textbf{w/o GRPO}: supervised training only;
(ii) \textbf{GRPO (Self M2T)}: the main setting, where the reward is computed by
re-inferring a pseudo caption with DiMo’s motion-to-text (M2T) branch and measuring
its alignment with the ground-truth text;
and (iii) \textbf{GRPO (Motion Extractor)}: computing the reward with a separately
trained motion--text feature extractor with similar setup as the
automatic evaluator used for reporting metrics.

As shown in Table~\ref{tab:grpo_reward_ablation}, both GRPO variants consistently
improve T2M FID over supervised training, while maintaining or slightly improving
M2T text quality. The Motion Extractor variant achieves the best FID, with M2T
performance closely matching the Self M2T variant. However, since the Motion
Extractor reward is trained with the similar setup of metric-reporting evaluator family,
using it as the default setting may blur the separation between optimization and
evaluation. Therefore, we keep \textbf{GRPO (Self M2T)} as the default setting in
the main paper and present \textbf{GRPO (Motion Extractor)} as an ablation.

Overall, these results verify that GRPO can be effectively applied to fine-tune
our unified model under different reward definitions. More broadly, an interesting
direction is to explore more independent reward modeling strategies---for example,
leveraging human preference signals collected from user studies---as a complementary
supervision source for post training fine-tuning.

\begin{table*}[t]
\centering
\caption{Ablation on GRPO reward design (HumanML3D).}
\label{tab:grpo_reward_ablation}
\small
\begin{tabular}{l|cccc|cccc}
\toprule
Method
& R@1 & R@2 & R@3 & FID$\downarrow$
& M2T R@1 & BLEU@1 & CIDEr & BERTScore \\
\midrule
Ours w/o GRPO
& 0.528 & 0.723 & 0.818 & 0.050
& 0.569 & 63.9 & 57.2 & 37.5 \\

Ours w/ GRPO (Self M2T)
& 0.528 & 0.724 & 0.818 & 0.047
& 0.577 & 64.2 & 58.1 & 37.7 \\

Ours w/ GRPO (Motion Extractor)
& 0.526 & 0.720 & 0.821 & \textbf{0.041}
& 0.574 & 64.0 & 57.6 & 37.7 \\
\bottomrule
\end{tabular}
\end{table*}

\subsection{Ablations on Other Datasets}
\label{app:motionx}
Since most previous unified model-based works did not use the Motion-X dataset,
direct comparisons are not available and Motion-X does not provide a unified and standardized public evaluation protocol. Therefore, we re-implement several representative text-to-motion baselines under a consistent experimental setup, including
T2M-GPT and MoMask, and evaluate 
for the T2M task only. The quantitative results are summarized in Table~\ref{tab:motionx_t2m}.

From the results, we observe that our method achieves performance on par with strong
T2M-specific models such as MoMask in retrieval metrics,
while obtaining the best FID score, indicating superior motion realism
and better alignment with the real-data distribution.

We also include a more complete coverage of prior work done on KIT-ML in Table~\ref{tab:kit_all}. As MotionGPT3 reported only results for model trained on single task for KIT-ML dataset, here we group them under separated model.

\begin{table}[!ht]
\centering
\caption{Motion-X results on the Text-to-Motion (T2M) task with reproduced baselines.
Due to the lack of a standardized evaluation protocol and missing M2T reports in prior work,
we only compare T2M performance under a unified setup.}
\label{tab:motionx_t2m}
\small
\begin{tabular}{l|cccccc}
\toprule
Method & R@1$\uparrow$ & R@2$\uparrow$ & R@3$\uparrow$ & FID$\downarrow$ & Div$\rightarrow$ & MM Dist$\downarrow$ \\
\midrule
Real motions & 0.510 & 0.691 & 0.791 & --    & 9.442 & 3.310 \\
T2M-GPT      & 0.370 & 0.546 & 0.654 & 2.174 & \textbf{9.303} & 4.252 \\
MoMask       & 0.287 & 0.445 & 0.554 & 0.884 & 8.400 & \textbf{2.792} \\
Ours         & \textbf{0.376} & \textbf{0.556} & \textbf{0.665} & \textbf{0.870} & 8.330 & 4.154 \\
\bottomrule
\end{tabular}
\end{table}

% ---------------- KIT-ML T2M&M2T full-----------------
 \begin{table*}[!ht]

\caption{Quantitative results of KIT-ML Text-to-Motion and Motion-to-Text. † marks as single-task
model reported by author \cite{zhu2025motiongpt3}.}
\vspace{-6pt}
\begin{center}
\label{tab:kit_all}
\setlength{\tabcolsep}{2.8pt}
\renewcommand{\arraystretch}{1.1}
\small
\resizebox{\textwidth}{!}{%
\begin{tabular}{clccccccc|ccccccc}
\toprule
\multirow{2}{*}{\centering Category} 
& \multirow{2}{*}{\centering Method} 
& \multicolumn{7}{c|}{\textbf{T2M}} 
& \multicolumn{7}{c}{\textbf{M2T}} \\

\cmidrule(lr){3-9}  
\cmidrule(l){10-16}

&& R@1$\uparrow$ & R@2$\uparrow$ & R@3$\uparrow$ & FID$\downarrow$ & Div$\rightarrow$ & MM$\uparrow$ & MM Dist$\downarrow$
& R@1$\uparrow$ & R@3$\uparrow$ & BLEU@1$\uparrow$ & BLEU@4$\uparrow$ & ROUGE-L$\uparrow$ & CIDEr$\uparrow$ & BERTScore$\uparrow$ \\

\midrule
   & Real Motion   & 0.424 & 0.649 & 0.779 & 0.031 & 11.08 & - & 2.788 & 0.399 & 0.793 & - & - & - & - & - \\
\midrule
\multirow{10}{*}{\centering T2M Only} 
 & MDM           & -     & -     & 0.396 & 0.497 & 10.847 & 1.907 & 9.191 & - & - & - & - & - & - & - \\
 & MotionDiffuse & 0.417 & 0.621 & 0.739 & 1.954 & 11.1  & 0.730 & 2.958 & - & - & - & - & - & - & - \\
 & MLD           & 0.39  & 0.609 & 0.734 & 0.404 & 10.8  & 2.192 & 3.204 & - & - & - & - & - & - & - \\
 & MoMask        & 0.433 & 0.656 & 0.781 & 0.204 & -     & 1.131 & 2.779 & - & - & - & - & - & - & - \\
 & T2M-GPT       & 0.416 & 0.627 & 0.745 & 0.514 & 10.921 & 1.570 & 3.007 & - & - & - & - & - & - & - \\
 & ReMoDiffuse   & 0.427 & 0.641 & 0.765 & 0.155 & 6.371  & 1.239 & 2.814 & - & - & - & - & - & - & - \\
 & DiverseMotion & 0.416 & 0.637 & 0.760 & 0.468 & 10.873 & 2.062 & 2.892 & - & - & - & - & - & - & - \\
 & MoGenTS       & 0.445 & 0.671 & 0.797 & 0.143 & 10.918 & -     & 2.711 & - & - & - & - & - & - & - \\
 & BAMM          & 0.438 & 0.661 & 0.788 & 0.183 & 11.008 & 1.609 & 2.723 & - & - & - & - & - & - & - \\
 & ReMoMask      & 0.453 & 0.682 & 0.805 & 0.138 & 10.830 & 2.017 & 2.682 & - & - & - & - & - & - & - \\
 & BAD          & 0.417 & 0.631 & 0.750 & 0.221 & 11.000 & 1.170 & 2.941 & - & - & - & - & - & - & - \\
 & MARDM        & 0.387 & 0.610 & 0.749 & 0.242 & - & 1.312 & 3.374 & - & - & - & - & - & - & - \\

\midrule
\multirow{3}{*}{\centering \shortstack{Separated\\Model}} 
 & TM2T          & 0.280 & 0.463 & 0.587 & 3.599 & 9.473  & 3.292 & 4.591 & 0.359 & 0.668 & 46.70 & 18.40 & 44.20 & 79.50 & 23.00 \\
 & LaMP          & 0.479 & 0.691 & 0.826 & 0.141 & 10.929 & -     & 2.704 & 0.540 & 0.844 & -    & -    & -    & -    & - \\
 & MotionGPT3†    & 0.456 & 0.680 & 0.803 & 0.227 & 11.026 & 0.904 & 2.704 & - & - & - & - & - & - & - \\

\midrule
\multirow{4}{*}{\centering \shortstack{Unified\\Model}} 
 & MotionGPT     & 0.366 & 0.558 & 0.680 & 0.510 & 10.350 & 2.328 & 3.527 & - & - & - & - & - & - & - \\
 & MotionGPT2    & \textbf{0.427} & 0.627 & \textbf{0.764} & 0.614 & \textbf{11.256} & 2.357 & 3.164 & - & - & - & - & - & - & - \\
 & MoTe          & 0.419 & 0.627 & 0.741 & 0.256 & -     & \textbf{2.615} & 3.216 & \textbf{0.421} & \textbf{0.765} & 44.90 & 14.51 & 41.80 & 55.60 & 35.90 \\
 & Ours w/o GRPO & 0.406 & 0.620 & 0.741 & \textbf{0.206} & 10.892 & 1.690 & \textbf{2.983} & 0.396 & 0.723 & \textbf{52.50} & \textbf{17.80} & \textbf{48.00} & \textbf{68.70} & \textbf{37.70} \\

\bottomrule
\end{tabular}%
}
\end{center}
\end{table*}

% \subsection{Results on KIT-ML}

% % ---------------- KIT-ML T2M&M2T-----------------
% \begin{table*}[h]
% \centering
% \caption{Quantitative results of Text-to-Motion and Motion-to-Text on \textbf{KIT-ML}}
% \label{tab:kitml_results}
% \setlength{\tabcolsep}{2.8pt}
% \renewcommand{\arraystretch}{1.1}
% \small
% \resizebox{\textwidth}{!}{%
% \begin{tabular}{lccccccc|ccccccc}
% \toprule
% % & \multicolumn{7}{c}{\textbf{T2M}}  \multicolumn{7}{c}{\textbf{M2T}} \\
% & \multicolumn{7}{c|}{\textbf{T2M}} & \multicolumn{7}{c}{\textbf{M2T}} \\
% \cmidrule(lr){2-8}  \cmidrule(l){9-15}
% Method
%  & R@1$\uparrow$ & R@2$\uparrow$ & R@3$\uparrow$ & FID$\downarrow$ & Div$\uparrow$ & MM$\uparrow$ & MM Dist$\downarrow$ & R@1$\uparrow$ & R@3$\uparrow$ & BLEU@1$\uparrow$ & BLEU@4$\uparrow$ & ROUGE-L$\uparrow$ & CIDEr$\uparrow$ & BERTScore$\uparrow$ \\
% \midrule
% TM2T          & 0.280 & 0.463 & 0.587 & 3.599 & 9.473 & 3.292 & 4.591 & 0.359 & 0.668 & 46.7 & 18.4 & 44.2 & 79.5 & 23.0 \\
% MotionGPT2    & 0.427 & 0.627 & 0.764 & 0.614 & 11.256 & 2.357 & 3.164 & - & - & - & - & - & - & - \\
% MoMask        & 0.433 & 0.656 & 0.781 & 0.204 & 10.711 & 1.131 & 2.779 & - & - & - & - & - & - & - \\
% MoTe          & 0.419 & 0.627 & 0.741 & 0.256 & - & 2.615 & 3.216 & 0.421 & 0.765 & 44.9 &  14.51 & 41.8 & 55.6 & 35.9 \\
% \midrule
% \makecell[l]{DiMo \\ (w/o GRPO)} &  &   &   &   &   &   &   &   &  &  &  &   &   &   \\
% \bottomrule
% \makecell[l]{DiMo \\ (with GRPO)} &   &   &   &   &   &   &   &   &  &  &  &   &   &   \\
% \bottomrule
% \end{tabular}%
% }
% \end{table*}

\subsection{Effect of Masking Schedule}
\label{app:mask_schedule}
\textbf{Effect of Masking Schedule.}
We further ablate the effect of masking schedules for text and motion tokens.
Table~\ref{tab:mask_schedule} compares the baseline configuration,
which applies a linear masking schedule to both modalities,
with two alternatives:
(i) applying a cosine schedule to motion tokens while keeping text linear
(\texttt{mcos\_linear}),
and (ii) applying cosine schedules to both text and motion tokens
(\texttt{mcos\_mcos}).

We observe that the baseline linear masking yields the best overall retrieval accuracy
and the lowest FID.
In contrast, cosine masking significantly degrades motion fidelity,
especially when applied jointly to both text and motion.
Interestingly, cosine-based schedules result in slightly higher diversity,
suggesting that they encourage more varied predictions at the cost of semantic consistency.
Overall, these results indicate that linear masking provides a more stable learning signal
for DiMo, and mixing masking schedules across modalities does not yield additional benefits.

\begin{table*}[t]
\centering
\caption{Ablation on masking schedules for text and motion tokens. (\texttt{mcos\_mcos}).}
\label{tab:mask_schedule}
\small
\begin{tabular}{l|ccccc}
\toprule
Schedule
& R@1$\uparrow$ & R@2$\uparrow$ & R@3$\uparrow$
& FID$\downarrow$ & Diversity$\rightarrow$ \\
\midrule
Baseline (linear + linear)
& \textbf{0.528} & \textbf{0.723} & \textbf{0.818}
& \textbf{0.050} & \textbf{9.300} \\
mcos\_linear (motion=cosine, text=linear)
& 0.502 & 0.699 & 0.795
& 0.369 & 9.030 \\
mcos\_mcos (motion=cosine, text=cosine)
& 0.507 & 0.701 & 0.798
& 0.264 & 8.947 \\
\bottomrule
\end{tabular}
\end{table*}

% ---------------- Ablation Motion-Text ratio-----------------

\subsection{Effect of Multi-task Proportion}
\label{sec:ablation_ratio}

Table~\ref{tab:ablation_ratio} examines how different training mixtures of text-to-motion (T2M), motion-to-text (M2T), and motion-only samples influence performance. 
We observe that M2T results are largely invariant to the data ratio: BLEU, ROUGE, CIDEr, and BERTScore remain stable even when the proportion of captioning samples is reduced. 
This indicates that the pretrained language backbone already provides strong priors for text generation, such that extensive exposure to captioning data is not required for competitive M2T performance. 

In contrast, T2M quality exhibits greater sensitivity to the task mixture. 
Allocating excessive capacity to M2T leads to a measurable decline in motion fidelity (lower recall, higher FID), reflecting the higher difficulty of modeling discrete motion tokens compared to text. 
The 8:1:1 configuration yields the most favorable balance, achieving strong language grounding while preserving the highest motion generation quality.

\begin{table*}[ht]
\centering
\caption{Ablation study: model performance under different data mixture ratios.}
\vspace{1em}
\label{tab:ablation_ratio}
\setlength{\tabcolsep}{2.8pt}
\renewcommand{\arraystretch}{1.1}
\small
\resizebox{\textwidth}{!}{%
\begin{tabular}{ccc|cccc|ccccccc}
\toprule
\makecell[c]{Text-to-Motion \\ (\%)} & \makecell[c]{Motion \\ (\%)} & \makecell[c]{Motion-to-Text \\ (\%)} 
& \multicolumn{4}{c|}{Text-to-Motion} & \multicolumn{7}{c}{Motion-to-Text} \\
\cmidrule(r){4-7} \cmidrule(r){8-14}
 & & & R@1$\uparrow$ & R@2$\uparrow$ & R@3$\uparrow$ & FID$\downarrow$ & R@1$\uparrow$ & R@3$\uparrow$ & BLEU@1$\uparrow$ & BLEU@4$\uparrow$ & ROUGE-L$\uparrow$ & CIDEr$\uparrow$ & BERTScore$\uparrow$\\
\midrule
80 & 10 & 10 & \textbf{0.528} & \textbf{0.723} & 0.818 & \textbf{0.050} 
        & 0.569 & \textbf{0.850} & 63.9 & \textbf{22.6} & 47.0 & 57.2 & 37.6 \\
70 & 10 & 20 & 0.523 & 0.722 & \textbf{0.820} & 0.091 
        & \textbf{0.578} & \textbf{0.850} & \textbf{64.4} & 22.3 & \textbf{47.1} & \textbf{58.2} & \textbf{38.1} \\
60 & 10 & 30 & 0.515 & 0.712 & 0.810 & 0.077 
        & 0.553 & 0.825 & 60.4 & 19.6 & 44.4 & 53.0 & 35.8 \\
\bottomrule
\end{tabular}
}
\end{table*}

% \section{More Qualitative Results}

\subsection{Effect of Classifier-Free Guidance (CFG) at Inference}
\label{sec:ablation_cfg}

We further study the impact of classifier-free guidance (CFG) scale on inference quality. 
Table~\ref{tab:cfg_ablation} reports results across scales $\{2,3,4,5,6\}$. 
We find that retrieval accuracy (R@1/2/3) improves when increasing the scale from 2 to 4, 
with $s=4$ yielding the highest recall. 
FID is lowest around $s=3$--$4$, while larger scales ($s \ge 5$) noticeably degrade fidelity. 
Diversity also reaches its maximum at $s=3$, whereas the matching score remains relatively stable across different settings. 

Overall, moderate CFG values ($s=3$--$4$) provide the best trade-off, enhancing semantic alignment without sacrificing motion quality, 
while overly strong guidance harms realism and reduces diversity.

\begin{table*}[h]
\centering
\caption{Ablation on classifier-free guidance (CFG) scale during inference. 
Moderate scales ($s=3$--$4$) achieve the best balance between motion fidelity and semantic alignment.}
\vspace{1em}
\label{tab:cfg_ablation}
\small
\begin{tabular}{c|ccc|c|c|c}
\toprule
CFG Scale & R@1$\uparrow$ & R@2$\uparrow$ & R@3$\uparrow$ & FID ↓ \\
\midrule
2 & 0.520 & 0.714 & 0.811 & 0.063  \\
3 (Ours) & \textbf{0.528} & 0.723 & \textbf{0.818} & \textbf{0.050} \\
4 & \textbf{0.528} & \textbf{0.725} & 0.817 & 0.059  \\
5 & 0.525 & 0.721 & \textbf{0.818} &0.067 \\
6 & 0.526 & 0.720 & 0.814 & 0.112 \\
\bottomrule
\end{tabular}
\end{table*}

\

\subsection{Effect of Down-weighting \texttt{[PAD]} Tokens}
\label{sec:pad_prob}

We found that the BERT backbone tends to over-generate \texttt{[PAD]} tokens at inference time. 
To mitigate this, we down-weight their probability by a multiplicative factor. 
Table~\ref{tab:pad_prob} compares different settings. 
A too aggressive penalty ($0.7$) harms caption quality, leading to lower BLEU, ROUGE, CIDEr, and BERTScore. 
In contrast, a mild penalty ($0.8$) achieves the best results, 
substantially improving caption fluency and semantic alignment while maintaining retrieval performance. 
This demonstrates that carefully tuning the \texttt{[PAD]} suppression factor is crucial for balancing sequence validity and output quality.

\begin{table}[t]
\centering

\caption{Ablation on down-weighting the probability of generating \texttt{[PAD]} tokens during text inference. 
We compare multiplicative factors of $0.7$ and $0.8$ against the baseline ($0.85$).}
\vspace{1em}
\label{tab:pad_prob}
\small
\begin{tabular}{c|cc|cc|ccc}
\toprule
Pad prob factor & R@1 & R@3 & BLEU@1 & BLEU@4 & ROUGE-L & CIDEr & BERTScore \\
\midrule
0.70 & 0.571 & 0.849 & 44.9 & 12.2 & 38.3 & 29.9 & 29.7 \\
0.80 & \textbf{0.585} & \textbf{0.861} & \textbf{61.7} & \textbf{20.6} & \textbf{46.0} & \textbf{56.3} & \textbf{37.1} \\
0.85 (baseline) & 0.558 & 0.842 & 60.4 & 19.5 & 44.4 & 52.9 & 35.8 \\
\bottomrule
\end{tabular}
\end{table}

\textit{Note:} Results reported in the appendix may vary slightly from the main tables due to differences in experimental runs and random seeds.  

\subsection{Effect of Multi-task Training (M2M / T2M / M2T)}

To study the effect of the multi-task formulation, we perform an ablation over the three training objectives:
(i) Motion-to-Motion (M2M),
(ii) Text-to-Motion (T2M), and
(iii) Motion-to-Text (M2T).
We toggle each objective on/off and report the full set of T2M and M2T metrics on HumanML3D. 
Results are summarized in Table~\ref{tab:multi_task_ablation}.
 
\begin{table*}[!ht]
\centering
\scriptsize
\setlength{\tabcolsep}{3pt}
\caption{
\textbf{Multi-task ablation on HumanML3D.}
Removing any single objective degrades either T2M or M2T performance,
while the full configuration (M2M + T2M + M2T) achieves the best overall trade-off across metrics.
}
\vspace{1em}
\label{tab:multi_task_ablation}

\resizebox{\textwidth}{!}{
\begin{tabular}{
c c c |
c c c |
c c c c |
c c |
c c c c c
}
\toprule
\multicolumn{3}{c|}{Tasks} &
\multicolumn{3}{c|}{T2M Retrieval $\uparrow$} &
\multicolumn{4}{c|}{T2M Generation} &
\multicolumn{2}{c|}{M2T Retrieval $\uparrow$} &
\multicolumn{5}{c}{M2T Text $\uparrow$} \\
M2M & T2M & M2T &
R@1 & R@2 & R@3 &
FID $\downarrow$ & Div $\rightarrow$ & MM $\uparrow$ & MM Dist $\downarrow$ &
R@1 & R@3 &
BLEU@1 & BLEU@4 & ROUGE-L & CIDEr & BERTScore \\
\midrule

No  & No  & Yes &
-- & -- & -- &
-- & -- & -- & -- &
0.547 & 0.822 &
56.9 & 16.3 & 42.8 & 45.9 & 34.3 \\

No  & Yes & No  &
0.518 & 0.704 & 0.801 &
0.323 & 9.335 & 2.014 & 2.999 &
-- & -- &
-- & -- & -- & -- & -- \\

No  & Yes & Yes &
\textbf{0.530} & \textbf{0.725} & \textbf{0.824} &
0.114 & 9.583 & \textbf{2.399} & \textbf{2.839} &
0.561 & 0.849 &
\textbf{65.3} & \textbf{23.2} & \textbf{47.2} & \textbf{57.6} & 37.4 \\

Yes & No  & Yes &
-- & -- & -- &
-- & -- & -- & -- &
0.559 & 0.838 &
56.4 & 15.9 & 42.5 & 47.1 & 34.5 \\

Yes & Yes & No &
0.515 & 0.714 & 0.814 &
0.174 & 9.146 & 2.106 & 2.956 &
-- & -- &
-- & -- & -- & -- & -- \\

Yes & Yes & Yes &
0.528 & 0.723 & 0.818 &
\textbf{0.050} & \textbf{9.515} & 2.016 & 2.867 &
\textbf{0.569} & \textbf{0.850} &
63.9 & 22.6 & 47.0 & 57.2 & \textbf{37.5} \\
\bottomrule
\end{tabular}
}
\end{table*}

\noindent
The results show that the three objectives are mutually beneficial. 
Training only with T2M improves the T2M branch but yields weaker motion semantics, 
while training only with M2T produces stronger captioning but degrades motion quality. 
Introducing M2M further enhances motion coherence and enables effective classifier-free guidance. 
The complete configuration (M2M + T2M + M2T) achieves the best balance on both T2M and M2T metrics, 
indicating that unified optimization encourages better motion–text alignment and motion structure.

\section{Limitations and Future Directions}
\label{app:limitations}

Despite the encouraging results, DiMo has several limitations that merit
discussion and point to important directions for future work.

\paragraph{Modeling Trade-offs.}
DiMo prioritizes perceptual realism and distributional fidelity in motion
generation, as reflected by strong FID scores. However, this design choice does
not always translate into peak performance on retrieval-based metrics such as
R-Precision, which are sensitive to the specific behavior of a fixed
text--motion evaluator. As discussed in Sec.~\ref{sec:experiments} and
Appendix~\ref{app:user_study}, optimizing for perceptual quality may lead to
generated motions that are closer to the true data distribution but less
aligned with the inductive biases of the evaluator. Understanding how to better
balance perceptual fidelity and evaluator-specific retrieval scores remains an
open problem.

\paragraph{Evaluation Protocol Limitations.}
Current T2M evaluation relies heavily on automatic retrieval-based metrics.
Our analysis shows that even ground-truth motions achieve relatively low
R@1 scores under the same evaluator, suggesting that these metrics favor
in-distribution motion patterns rather than fully capturing semantic alignment
as perceived by humans. While we complement quantitative results with a human
preference study, designing more robust and interpretable evaluation protocols
for text--motion alignment remains an important challenge for the community.

\paragraph{Reward Design.}
We adopt GRPO to improve alignment and controllability using task-specific rewards
that are lightweight and model-aware. Our ablations (Appendix~\ref{app:reward_ablation})
show that GRPO consistently improves T2M FID while maintaining strong M2T performance.
More broadly, reward design involves practical trade-offs between simplicity,
generality, and the type of supervision signal available. A potentially stronger
alternative is to train a dedicated reward model calibrated to human judgments
(e.g., from user-study preference data) and use it as a complementary signal for
motion--language alignment finetuning.

Overall, these limitations highlight that unified text--motion modeling
involves non-trivial trade-offs between generation quality, semantic alignment,
and evaluation methodology. We hope DiMo serves as a step toward more principled
and holistic solutions in this space.

%% file: icml2026.bib
@article{rose1998verbs,
  title={Verbs and adverbs: Multidimensional motion interpolation},
  author={Rose, Charles and Cohen, Michael F and Bodenheimer, Bobby},
  journal={IEEE Computer Graphics and Applications},
  volume={18},
  number={5},
  pages={32--40},
  year={1998},
  publisher={IEEE}
}

@incollection{mukai2005geostatistical,
  title={Geostatistical motion interpolation},
  author={Mukai, Tomohiko and Kuriyama, Shigeru},
  booktitle={ACM SIGGRAPH 2005 Papers},
  pages={1062--1070},
  year={2005}
}

@inproceedings{barsoum2018hp,
  title={Hp-gan: Probabilistic 3d human motion prediction via gan},
  author={Barsoum, Emad and Kender, John and Liu, Zicheng},
  booktitle={Proceedings of the IEEE conference on computer vision and pattern recognition workshops},
  pages={1418--1427},
  year={2018}
}

@article{harvey2020robust,
  title={Robust motion in-betweening},
  author={Harvey, F{\'e}lix G and Yurick, Mike and Nowrouzezahrai, Derek and Pal, Christopher},
  journal={ACM Transactions on Graphics (TOG)},
  volume={39},
  number={4},
  pages={60--1},
  year={2020},
  publisher={ACM New York, NY, USA}
}

@inproceedings{aliakbarian2020stochastic,
  title={A stochastic conditioning scheme for diverse human motion prediction},
  author={Aliakbarian, Sadegh and Saleh, Fatemeh Sadat and Salzmann, Mathieu and Petersson, Lars and Gould, Stephen},
  booktitle={Proceedings of the IEEE/CVF Conference on Computer Vision and Pattern Recognition},
  pages={5223--5232},
  year={2020}
}

@inproceedings{petrovich2021action,
  title={Action-conditioned 3d human motion synthesis with transformer vae},
  author={Petrovich, Mathis and Black, Michael J and Varol, G{\"u}l},
  booktitle={Proceedings of the IEEE/CVF International Conference on Computer Vision},
  pages={10985--10995},
  year={2021}
}

@inproceedings{tevet2022human,
  title={Human Motion Diffusion Model},
  author={Tevet, Guy and Raab, Sigal and Gordon, Brian and Shafir, Yoni and Cohen-or, Daniel and Bermano, Amit Haim},
  booktitle={The Eleventh International Conference on Learning Representations},
  year={2022}
}

@article{zhang2024motiondiffuse,
  title={Motiondiffuse: Text-driven human motion generation with diffusion model},
  author={Zhang, Mingyuan and Cai, Zhongang and Pan, Liang and Hong, Fangzhou and Guo, Xinying and Yang, Lei and Liu, Ziwei},
  journal={IEEE Transactions on Pattern Analysis and Machine Intelligence},
  year={2024},
  publisher={IEEE}
}

@inproceedings{karunratanakul2023guided,
  title={Guided motion diffusion for controllable human motion synthesis},
  author={Karunratanakul, Korrawe and Preechakul, Konpat and Suwajanakorn, Supasorn and Tang, Siyu},
  booktitle={Proceedings of the IEEE/CVF International Conference on Computer Vision},
  pages={2151--2162},
  year={2023}
}

@article{zhang2024finemogen,
  title={Finemogen: Fine-grained spatio-temporal motion generation and editing},
  author={Zhang, Mingyuan and Li, Huirong and Cai, Zhongang and Ren, Jiawei and Yang, Lei and Liu, Ziwei},
  journal={Advances in Neural Information Processing Systems},
  volume={36},
  year={2024}
}

@inproceedings{gong2023tm2d,
  title={Tm2d: Bimodality driven 3d dance generation via music-text integration},
  author={Gong, Kehong and Lian, Dongze and Chang, Heng and Guo, Chuan and Jiang, Zihang and Zuo, Xinxin and Mi, Michael Bi and Wang, Xinchao},
  booktitle={Proceedings of the IEEE/CVF International Conference on Computer Vision},
  pages={9942--9952},
  year={2023}
}

@article{zhang2024large,
      title   =   {Large Motion Model for Unified Multi-Modal Motion Generation}, 
      author  =   {Zhang, Mingyuan and Jin, Daisheng and Gu, Chenyang and Hong, Fangzhou and Cai, Zhongang and Huang, Jingfang and Zhang, Chongzhi and Guo, Xinying and Yang, Lei and He, Ying and Liu, Ziwei},
      year    =   {2024},
      journal =   {arXiv preprint arXiv:2404.01284},
}

@inproceedings{guo2022generating,
  title={Generating Diverse and Natural 3D Human Motions From Text},
  author={Guo, Chuan and Zou, Shihao and Zuo, Xinxin and Wang, Sen and Ji, Wei and Li, Xingyu and Cheng, Li},
  booktitle={Proceedings of the IEEE/CVF Conference on Computer Vision and Pattern Recognition},
  pages={5152--5161},
  year={2022}
}

@inproceedings{petrovich2022temos,
  title={TEMOS: Generating diverse human motions from textual descriptions},
  author={Petrovich, Mathis and Black, Michael J and Varol, G{\"u}l},
  booktitle={European Conference on Computer Vision},
  pages={480--497},
  year={2022},
  organization={Springer}
}

@article{jiang2024motiongpt,
  title={Motiongpt: Human motion as a foreign language},
  author={Jiang, Biao and Chen, Xin and Liu, Wen and Yu, Jingyi and Yu, Gang and Chen, Tao},
  journal={Advances in Neural Information Processing Systems},
  volume={36},
  year={2024}
}

@misc{wang2024motiongpt2,
      title={MotionGPT-2: A General-Purpose Motion-Language Model for Motion Generation and Understanding}, 
      author={Yuan Wang and Di Huang and Yaqi Zhang and Wanli Ouyang and Jile Jiao and Xuetao Feng and Yan Zhou and Pengfei Wan and Shixiang Tang and Dan Xu},
      year={2024},
      eprint={2410.21747},
      archivePrefix={arXiv},
      primaryClass={cs.CV},
 
}

@inproceedings{guo2022tm2t,
  title={Tm2t: Stochastic and tokenized modeling for the reciprocal generation of 3d human motions and texts},
  author={Guo, Chuan and Zuo, Xinxin and Wang, Sen and Cheng, Li},
  booktitle={European Conference on Computer Vision},
  pages={580--597},
  year={2022},
  organization={Springer}
}

@inproceedings{radford2018improving,
  title={Improving language understanding by generative pre-training},
  author={Radford, Alec and Narasimhan, Karthik and Salimans, Tim and Sutskever, Ilya},
  booktitle={OpenAI Technical Report},
  year={2018}
}

@inproceedings{brown2020language,
 author = {Brown, Tom and Mann, Benjamin and Ryder, Nick and Subbiah, Melanie and Kaplan, Jared D and Dhariwal, Prafulla and Neelakantan, Arvind and Shyam, Pranav and Sastry, Girish and Askell, Amanda and Agarwal, Sandhini and Herbert-Voss, Ariel and Krueger, Gretchen and Henighan, Tom and Child, Rewon and Ramesh, Aditya and Ziegler, Daniel and Wu, Jeffrey and Winter, Clemens and Hesse, Chris and Chen, Mark and Sigler, Eric and Litwin, Mateusz and Gray, Scott and Chess, Benjamin and Clark, Jack and Berner, Christopher and McCandlish, Sam and Radford, Alec and Sutskever, Ilya and Amodei, Dario},
 booktitle = {Advances in Neural Information Processing Systems},
 editor = {H. Larochelle and M. Ranzato and R. Hadsell and M.F. Balcan and H. Lin},
 pages = {1877--1901},
 publisher = {Curran Associates, Inc.},
 title = {Language Models are Few-Shot Learners},
 volume = {33},
 year = {2020}
}

@inproceedings{austin2021structured,
 author = {Austin, Jacob and Johnson, Daniel D. and Ho, Jonathan and Tarlow, Daniel and van den Berg, Rianne},
 booktitle = {Advances in Neural Information Processing Systems},
 editor = {M. Ranzato and A. Beygelzimer and Y. Dauphin and P.S. Liang and J. Wortman Vaughan},
 pages = {17981--17993},
 publisher = {Curran Associates, Inc.},
 title = {Structured Denoising Diffusion Models in Discrete State-Spaces},
 volume = {34},
 year = {2021}
}

@inproceedings{devlin2018bert,
    title = "{BERT}: Pre-training of Deep Bidirectional Transformers for Language Understanding",
    author = "Devlin, Jacob  and
      Chang, Ming-Wei  and
      Lee, Kenton  and
      Toutanova, Kristina",
    editor = "Burstein, Jill  and
      Doran, Christy  and
      Solorio, Thamar",
    booktitle = "Proceedings of the 2019 Conference of the North {A}merican Chapter of the Association for Computational Linguistics: Human Language Technologies, Volume 1 (Long and Short Papers)",
    month = jun,
    year = "2019",
    address = "Minneapolis, Minnesota",
    publisher = "Association for Computational Linguistics",
    doi = "10.18653/v1/N19-1423",
    pages = "4171--4186"
}

@inproceedings{li2022diffusionlm,
 author = {Li, Xiang and Thickstun, John and Gulrajani, Ishaan and Liang, Percy S and Hashimoto, Tatsunori B},
 booktitle = {Advances in Neural Information Processing Systems},
 editor = {S. Koyejo and S. Mohamed and A. Agarwal and D. Belgrave and K. Cho and A. Oh},
 pages = {4328--4343},
 publisher = {Curran Associates, Inc.},
 title = {Diffusion-LM Improves Controllable Text Generation},
 volume = {35},
 year = {2022}
}

@InProceedings{chang2022maskgit,
  title = {MaskGIT: Masked Generative Image Transformer},
  author={Huiwen Chang and Han Zhang and Lu Jiang and Ce Liu and William T. Freeman},
  booktitle = {The IEEE Conference on Computer Vision and Pattern Recognition (CVPR)},
  month = {June},
  year = {2022}
}

@article{yu2025discrete,
  title={Discrete Diffusion in Large Language and Multimodal Models: A Survey},
  author={Yu, Runpeng and Li, Qi and Wang, Xinchao},
  journal={arXiv preprint arXiv:2506.13759},
  year={2025}
}

@inproceedings{guo2023momask,
  author       = {Chuan Guo and
                  Yuxuan Mu and
                  Muhammad Gohar Javed and
                  Sen Wang and
                  Li Cheng},
  title        = {{MoMask}: Generative Masked Modeling of 3{D} Human Motions},
  booktitle    = {The IEEE Conference on Computer Vision and Pattern Recognition (CVPR)},
  year         = {2024}
}

@misc{shao2024deepseekmath,
  author = {Zhihong Shao and Peiyi Wang and Qihao Zhu and Runxin Xu and Junxiao Song and Mingchuan Zhang and Y.K. Li and Y. Wu and Daya Guo},
  title = {DeepSeekMath: Pushing the Limits of Mathematical Reasoning in Open Language Models},
  journal = {CoRR},
  volume = {abs/2402.03300},
  year = {2024},
}

@misc{zhu2025motiongpt3,
    title={MotionGPT3: Human Motion as a Second Modality}, 
    author={Bingfan Zhu and Biao Jiang and Sunyi Wang and Shixiang Tang and Tao Chen and Linjie Luo and Youyi Zheng and Xin Chen},
    year={2025},
    eprint={2506.24086},
    archivePrefix={arXiv},
    primaryClass={cs.CV},
}

@article{nie2025llada,
  title={Large Language Diffusion Models},
  author={Nie, Shen and Zhu, Fengqi and You, Zebin and Zhang, Xiaolu and Ou, Jingyang and Hu, Jun and Zhou, Jun and Lin, Yankai and Wen, Ji-Rong and Li, Chongxuan},
  journal={arXiv preprint arXiv:2502.09992},
  year={2025}
}

@misc{zeghidour2021soundstream,
    title   = {SoundStream: An End-to-End Neural Audio Codec},
    author  = {Neil Zeghidour and Alejandro Luebs and Ahmed Omran and Jan Skoglund and Marco Tagliasacchi},
    year    = {2021},
    eprint  = {2107.03312},
    archivePrefix = {arXiv},
    primaryClass = {cs.SD}
}

@inproceedings{DBLP:conf/iclr/LoshchilovH19,
  author       = {Ilya Loshchilov and
                  Frank Hutter},
  title        = {Decoupled Weight Decay Regularization},
  booktitle    = {7th International Conference on Learning Representations, {ICLR} 2019,
                  New Orleans, LA, USA, May 6-9, 2019},
  publisher    = {OpenReview.net},
  year         = {2019},
  bibsource    = {dblp computer science bibliography, https://dblp.org}
}

@article{zhang2019bertscore,
  title={Bertscore: Evaluating text generation with bert},
  author={Zhang, Tianyi and Kishore, Varsha and Wu, Felix and Weinberger, Kilian Q and Artzi, Yoav},
  journal={arXiv preprint arXiv:1904.09675},
  year={2019}
}

@inproceedings{vedantam2015cider,
  title={Cider: Consensus-based image description evaluation},
  author={Vedantam, Ramakrishna and Lawrence Zitnick, C and Parikh, Devi},
  booktitle={Proceedings of the IEEE conference on computer vision and pattern recognition},
  pages={4566--4575},
  year={2015}
}

@inproceedings{lin2004rouge,
  title={Rouge: A package for automatic evaluation of summaries},
  author={Lin, Chin-Yew},
  booktitle={Text summarization branches out},
  pages={74--81},
  year={2004}
}

@inproceedings{papineni2002bleu,
  title={Bleu: a method for automatic evaluation of machine translation},
  author={Papineni, Kishore and Roukos, Salim and Ward, Todd and Zhu, Wei-Jing},
  booktitle={Proceedings of the 40th annual meeting of the Association for Computational Linguistics},
  pages={311--318},
  year={2002}
}
